\pdftrailerid{}
\documentclass{article} 

\usepackage{template/fancyhdr,template/natbib,template/iclr2027_conference_preprint,times}

\usepackage{amsmath,amsfonts,bm}

\def\eqref#1{equation~\ref{#1}}

\def\1{\bm{1}}

\DeclareMathAlphabet{\mathsfit}{\encodingdefault}{\sfdefault}{m}{sl}
\SetMathAlphabet{\mathsfit}{bold}{\encodingdefault}{\sfdefault}{bx}{n}

\newcommand{\KL}{D_{\mathrm{KL}}}

\usepackage{booktabs}
\usepackage{array}
\usepackage{graphicx}
\usepackage{microtype}
\usepackage{multirow}
\usepackage[table,dvipsnames]{xcolor}

\usepackage{enumitem}
\usepackage{xspace}
\usepackage{pifont}
\usepackage{subcaption} 

\usepackage{siunitx}
\usepackage{float}
\usepackage{wrapfig}
\usepackage{hyperref}
\hypersetup{
    colorlinks,
    linkcolor={MidnightBlue},
    citecolor={MidnightBlue},
    urlcolor={MidnightBlue},
}
\usepackage{url}
\usepackage{subcaption}
\usepackage[capitalise,nameinlink,sort&compress]{cleveref}
\crefname{section}{Sec.}{Secs.}
\Crefname{section}{Section}{Sections}
\crefname{subsection}{Sec.}{Secs.}
\Crefname{subsection}{Section}{Sections}
\crefname{table}{Tab.}{Tabs.}
\Crefname{table}{Table}{Tables}
\crefname{figure}{Fig.}{Figs.}
\Crefname{figure}{Figure}{Figures}
\crefname{equation}{Eq.}{Eqs.}
\Crefname{equation}{Equation}{Equations}
\crefname{appendix}{App.}{Apps.}
\Crefname{appendix}{Appendix}{Appendices}

\DeclareMathOperator{\TC}{TC}
\newcommand{\KLargs}[2]{D_{\mathrm{KL}}\left(#1 \mathbin{\Vert} #2\right)}

\newcommand{\eg}{e.g.\xspace}
\newcommand{\ie}{i.e.\xspace}

\newcommand{\methodNormal}{JEM\xspace}
\newcommand{\method}{\textsc{\methodNormal}\xspace}
\newcommand{\dataset}{A140M\xspace}

\title{Scalable patch-level self-supervised learning}

\author{\begin{tabular*}{\textwidth}[t]{@{\extracolsep{\fill}}lccr@{}}
        Maximilian Seitzer & Gabriele Trivigno & Antonín Vobecký & Seungeun Yi\\
        \textbf{Maxime Oquab} & \textbf{Huy V. Vo} & \textbf{Oriane Siméoni} & \textbf{Piotr Bojanowski}
    \end{tabular*}\\
    \vspace{-0.65em}Meta FAIR \\
    \texttt{seitzer@meta.com}
}

\iclrfinalcopy \newcount\PaperFigureRoundingIndex
\newcommand{\PaperFigureRoundingX}[3]{\rlap{\kern-#1sp\raisebox{-#2sp}[0pt][0pt]{\hbox{\PaperFigureRoundingIndex=#3\relax
  \pdfliteral{q Q}\loop\ifnum\PaperFigureRoundingIndex>0\relax
  \kern-132sp\pdfliteral{q Q}\advance\PaperFigureRoundingIndex by-1\relax
  \repeat\kern\dimexpr132sp*#3\relax\pdfliteral{q Q}}}}}
\newcommand{\PaperFigureRoundingY}[3]{\rlap{\kern-#1sp\raisebox{-#2sp}[0pt][0pt]{\hbox{\PaperFigureRoundingIndex=0\relax
  \pdfliteral{q Q}\loop\ifnum\PaperFigureRoundingIndex<#3\relax
  \advance\PaperFigureRoundingIndex by1\relax
  \raisebox{\dimexpr132sp*\PaperFigureRoundingIndex\relax}[0pt][0pt]{\hbox{\pdfliteral{q Q}}}\repeat\pdfliteral{q Q}}}}}
\expandafter\def\csname PaperFigureRounding@figures/introduction/class-seg-comparison.pdf\endcsname{\PaperFigureRoundingY{32546317}{39077173}{20}\pdfliteral direct{1 0 0 1 0 16384 cm 1 0 0 1 0 -16384 cm}}
\expandafter\def\csname PaperFigureRounding@figures/scaling-flops-averages.pdf\endcsname{\PaperFigureRoundingX{33169097}{37970205}{22}}

\newsavebox{\PaperFigureBox}
\newcommand{\PaperFigureBoxContents}[4]{\ifnum#4=0\relax
    \sbox{\PaperFigureBox}{\includegraphics{#1}}\else
    \sbox{\PaperFigureBox}{\kern-#4bp\raisebox{-#4bp}{\includegraphics{#1}}\kern-#4bp}\fi
  \wd\PaperFigureBox=#2\relax
  \ht\PaperFigureBox=#3\relax
  \dp\PaperFigureBox=0pt
  \leavevmode\usebox{\PaperFigureBox}}
\newcommand{\DeclarePaperFigure}[7]{\expandafter\def\csname PaperFigure@#1\endcsname{\ifdim#7=0pt\else\hspace{#7}\fi
    \PaperFigureBoxContents{#1}{#2}{#3}{#4}\ifdim#5pt=0pt\ifdim#6pt=0pt\else
      \pdfliteral direct{1 0 0 1 #5 #6 cm}\fi\else
      \pdfliteral direct{1 0 0 1 #5 #6 cm}\fi
    \ifcsname PaperFigureRounding@#1\endcsname\csname PaperFigureRounding@#1\endcsname\fi
  }}
\newcommand{\includefigurepdf}[1]{\csname PaperFigure@#1\endcsname}
\DeclarePaperFigure{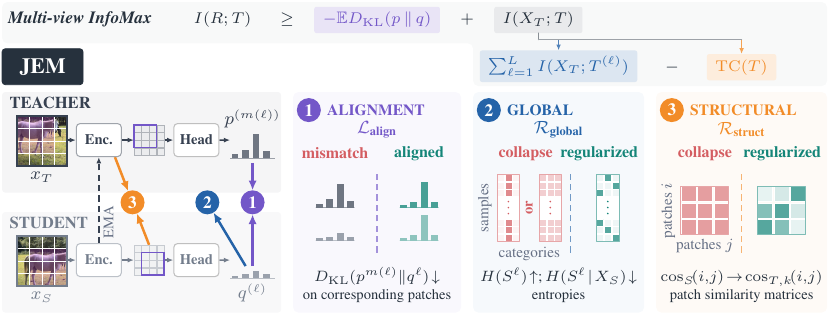}{26049573sp}{9993566sp}{1}{0}{0.000}{0pt}
\DeclarePaperFigure{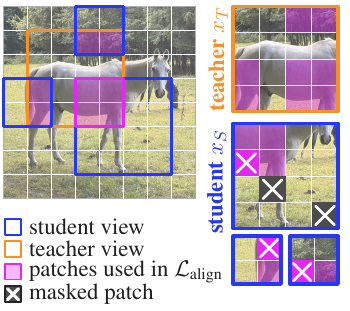}{10893472sp}{9662064sp}{1}{0.000}{0.00}{12.5pt}
\DeclarePaperFigure{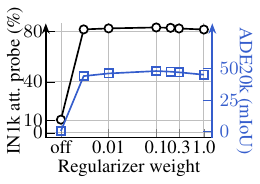}{8191650sp}{5800379sp}{0}{-0.001}{0.000}{0pt}
\DeclarePaperFigure{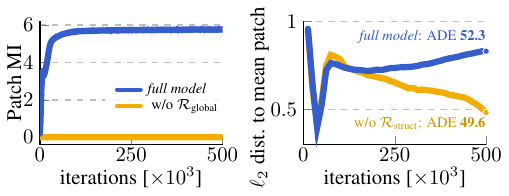}{16059661sp}{6158278sp}{0}{0.000}{0.001}{0pt}
\DeclarePaperFigure{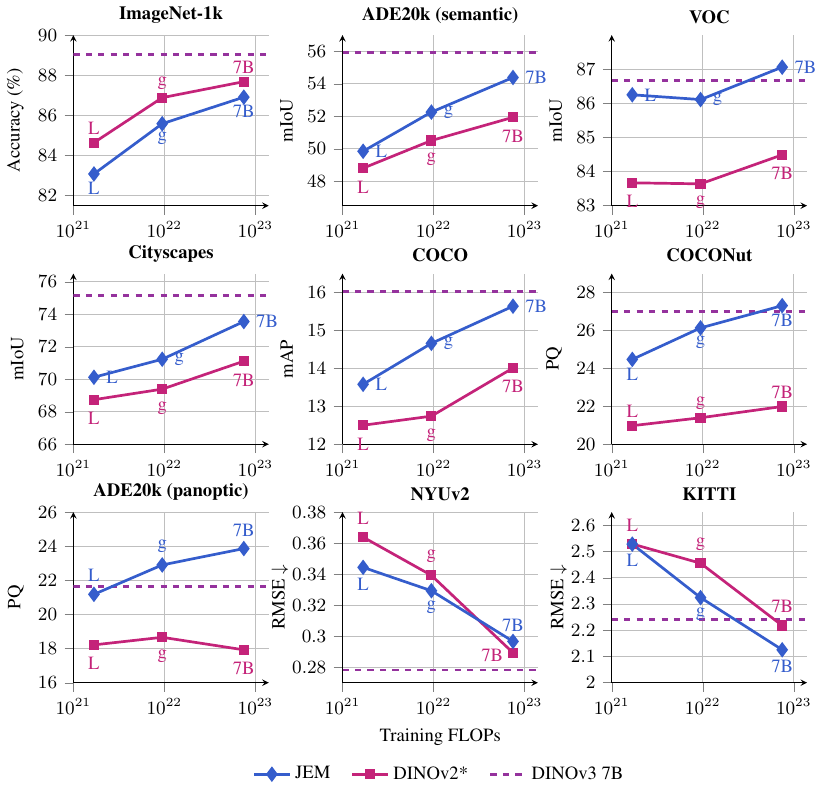}{25995687sp}{25176346sp}{0}{0.000}{-0.001}{0pt}
\DeclarePaperFigure{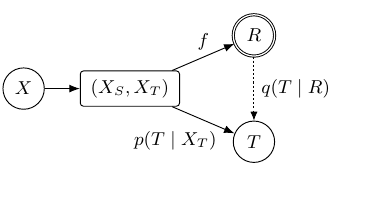}{11561004sp}{6246685sp}{0}{0.000}{0.000}{0pt}
\DeclarePaperFigure{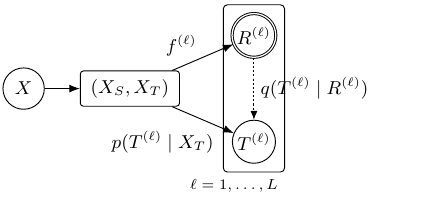}{13612162sp}{6246685sp}{0}{0.000}{0.000}{0pt}
\DeclarePaperFigure{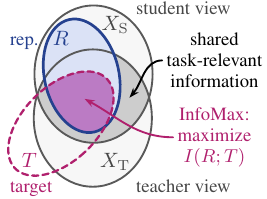}{8411179sp}{6262273sp}{0}{0.000}{0.000}{0pt}
\DeclarePaperFigure{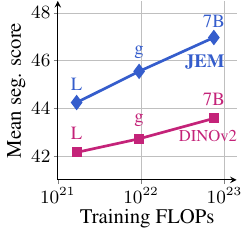}{7864641sp}{7410070sp}{0}{0.000}{0.000}{0pt}
\DeclarePaperFigure{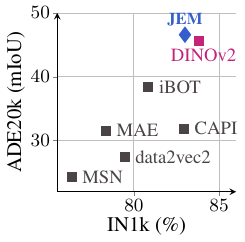}{7708711sp}{7636970sp}{0}{0.0009765625}{0.000}{0pt}
\DeclarePaperFigure{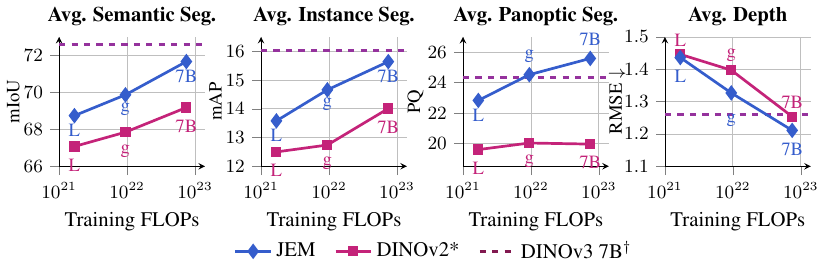}{26064668sp}{8743938sp}{0}{0.000}{0.000}{0pt}

\begin{document}

\maketitle
\vspace{-1em}
\begin{abstract}
Self-supervised learning (SSL) at scale produces powerful visual representations.
However, most scalable SSL methods rely on ad hoc combinations of multiple objectives and stabilization mechanisms.
Taking a step back, we ask if we can design a high-performing, yet principled SSL algorithm. 
Starting from the multi-view assumption---stipulating that task-relevant content is captured by the information common to different views---we construct an information-theoretic objective decomposing into interpretable terms.
This derivation yields \method, a student-teacher method that learns by aligning corresponding patch representations across views, explicitly regularized by information and structure preservation losses.
\method trains stably from 300M to 7B parameters, and, to our knowledge, is the first latent-space patch-level method demonstrated at 7B scale. 
Across all scales, \method reaches strong performance on both global and dense probing tasks, on segmentation benchmarks consistently surpassing the DINOv2 algorithm---an influential foundation for today's strongest visual SSL methods.
Notably, at 7B parameters, it exceeds the performance of DINOv3 on panoptic segmentation, despite being trained on 12$\times$ less data without refinement stages.
These results demonstrate that we can indeed design an SSL algorithm that learns strong representations, is principled and stable.

\end{abstract}

\vspace{-1em}
\section{Introduction}
\label{sec:intro}

Self-supervised learning has become a standard route to general-purpose visual representations, and the strongest encoders now match or exceed supervised pretraining across a wide range of tasks \citep{oquab2024dinov2,simeoni2025dinov3}. 
The best methods combine both image- and patch-level objectives, and have been successfully scaled \citep{oquab2024dinov2, simeoni2025dinov3, fan2025webssl,fu2026lingbotvision}.
However, this success comes at a price: these methods have grown complex, their objectives do not always compose naturally, and additional stabilization tricks are required to keep them from degrading each other~\citep{oquab2024dinov2,simeoni2025dinov3}. 
Impressive as these results are, we take them as an invitation to take a step back and reconsider what a principled self-supervised objective could look like.

\begin{figure}[h!]
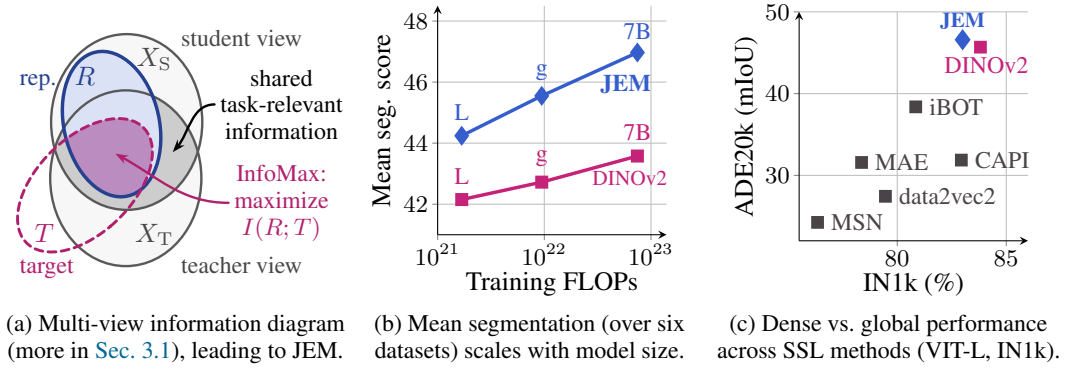
 \centering
    \begin{minipage}[b]{0.32\linewidth}
        \raisebox{6pt}[\height][\depth]{\includefigurepdf{figures/introduction/shared_information.pdf}\space}
        \subcaption{\centering 
        Multi-view information diagram (more in \cref{sec:info-framework}),
leading to JEM.
        }
        \label{fig:intro:info}
    \end{minipage}
    \hfill
    \begin{minipage}[b]{0.33\linewidth}
        \centering
        \includefigurepdf{figures/introduction/scaling-flops-intro.pdf}\space\space
        \vspace{-3pt}
        \subcaption{\centering Mean segmentation (over six datasets) scales with model size.}
        \label{fig:intro:scale}
    \end{minipage}
    \hfill
    \begin{minipage}[b]{0.33\linewidth}
        \centering
        \includefigurepdf{figures/introduction/class-seg-comparison.pdf}\space\space
        \vspace{-3pt}
        \subcaption{\centering Dense vs.\ global performance across SSL methods (VIT-L, IN1k).}
        \label{fig:intro:comparison}
    \end{minipage}
    \caption{
    \textbf{We introduce \method}, a student-teacher SSL objective grounded in information theory. 
    Following the multi-view principle (a), it maximizes the information a student representation retains about what views share. 
    Average scores over semantic, panoptic and instance segmentation at different scales (b) show that \method scales effectively and consistently outperforms DINOv2.  
    (c) \method surpasses methods using patch- and/or image-level objectives on dense tasks while remaining competitive on classification. 
    \method learns strong representations, is principled and stable.
    }
    \label{fig:first}
\end{figure}

Many successful methods are built from the idea that different views (altered through augmentations, cropping, masking) of the same image should be described similarly~\citep{chen2020simple,caron2021emerging,zbontar2021barlowtwins}. 
We believe that the underlying \emph{multi-view assumption}---which suggests that if each view is sufficient for the downstream task, then a representation should retain the information shared between views (\cref{fig:intro:info})---can act as a core learning principle.
From the multi-view assumption, we derive a multi-view InfoMax objective for student-teacher SSL with two terms: one maximizing cross-view alignment, one preventing information collapse.

We then discuss how to instantiate this objective into \method, a novel joint-embedding multi-view method trained with a simple patch-level cross-view alignment loss, and two regularization terms, one preventing global collapse, one preserving the spatial structure of the features.
We show that \method trains stably and scales well (\cref{fig:intro:scale}), without additional machinery, to model sizes up to 7B parameters.
\method performs strongly on dense benchmarks, consistently surpassing a DINOv2 baseline on semantic, instance, and panoptic segmentation, while remaining competitive with the strongest models on image-level classification (\cref{fig:intro:comparison}).
At 7B scale, \method even exceeds DINOv3's performance on panoptic segmentation, despite being trained for a shorter time on a smaller dataset without additional refinement steps.

To summarize, we 
(1) derive a principled information-theoretic SSL objective leading to \method, a joint-embedding multi-view method trained with a patch-level objective, that 
(2) trains stably across scales up to 7B parameters and
(3) outperforms the DINOv2 algorithm in controlled comparisons on dense prediction tasks while remaining competitive on classification, even exceeding the DINOv3 7B foundation model on panoptic segmentation.

\section{Related Work}
\paragraph{Self-Supervised Visual Representation Learning} 

Modern SSL falls into four broad families.
Contrastive methods \citep{he2020moco,chen2020simple} pull together representations of different views of the same image while pushing apart representations of other images.
Redundancy-reduction methods \citep{zbontar2021barlowtwins,bardes2021vicreg} enforce view invariance while explicitly preventing collapse across dimensions.
Self-clustering approaches such as SwAV~\cite{caron2021swav} and DINO~\citep{caron2021emerging} learn cluster assignments and align those across views.
Masked prediction instead removes part of the input and predicts the missing information: MAE~\citep{he2021masked} predicts pixels, while I-JEPA~\citep{assran2023self} predicts latent representations of spatially localized target blocks, and iBOT~\citep{zhou2021ibot} and CAPI~\citep{darcet2025capi} predict cluster assignments for masked patches.

Recent methods increasingly differ in whether they rely on a single objective family or combine complementary objectives.
I-JEPA and CAPI are examples of single-family latent prediction approaches: I-JEPA predicts latent representations of masked target blocks, while CAPI predicts latent cluster assignments for masked patches.
DINOv2~\citep{oquab2024dinov2}, in contrast, combines global DINO self-distillation with patch-level iBOT prediction and KoLeo~\citep{sablayrolles2018spreading} regularization; DINOv3~\citep{simeoni2025dinov3} even adds a further term, Gram anchoring, to
counter an effect observed during long training at large scale: global performance keeps improving while dense feature maps progressively lose spatial locality.
Are such combinations necessary for scaling SSL?
Our work explores a principled information-theoretic objective derived from maximizing mutual information between representations of different views.
Unlike methods with separate global and local objectives, our model has no class token and applies the objective exclusively to patch representations, while still matching the performance of multi-objective methods such as DINOv2.

\paragraph{Information theory in SSL}
Information-theoretic SSL spans several closely related perspectives~\citep{shwartzziv2023compress}.
The original InfoMax principle~\citep{linsker1988infomax} prescribes maximizing the information the representation retains about its input, implemented for representation learning by Deep InfoMax~\citep{hjelm2019learning}.
Other works instead maximize the information between different \emph{views} of the data~\citep{oord2018cpc,bachman2019learning,tian2020contrastive,henaff2020data}, yielding the \emph{multi-view InfoMax} principle; 
we also use it as our starting point.
However, instead of estimating mutual information with contrastive objectives or separate critics, we use a variational approximation which naturally fits the student-teacher paradigm common in modern SSL. Complementarily, the information bottleneck principle suggests retaining predictive information while compressing superfluous information~\citep{tishby1999information,alemi2017deep}.
\citet{federici2020learning} combine it with multi-view InfoMax by preserving shared and suppressing view-specific information.
In contrast, we derive an explicit anti-collapse term that \emph{maintains} view-specific information; however, our use of discrete representation targets can be seen as an architectural bottleneck.
The limits of mutual information as an objective for SSL have been discussed by~\citet{tschannen2020onmutualinfo,tian2020goodviews,tsai2021selfsupervised}.
For us, the multi-view InfoMax objective informs practical loss design, acknowledging that learning good representations also relies on inductive biases from architecture and view construction.
We review more related work in \cref{app:sec:extended-related-work}.

\section{Method}

We aim to build an SSL method that is \emph{principled} and \emph{stable}, without any sacrifice in \emph{downstream performance}.
Starting from the multi-view assumption, we derive an information-theoretic objective (\cref{sec:info-framework}). 
From this objective, we derive an alignment loss (\cref{subsec:alignemnt-loss}) and regularization terms that mitigate its collapse modes (\cref{subsec:global-collapse-prevention,subsec:struct-collapse-prevention}).
In \cref{subsec:view}, we then discuss the careful design of the views, which we found to be crucial for best performance.
Throughout the section we provide detailed design choices that make \method, a new joint-embedding SSL method yielding results competitive with more complex and highly tuned SSL objectives.

\subsection{The Multi-View InfoMax Objective}
\label{sec:info-framework}

The \emph{multi-view assumption} stipulates that any view of the data alone contains sufficient information for the downstream tasks~\citep{sridharan2008information}.
It follows that all task-relevant information must be contained in the \emph{shared information} across views.
Whether the multi-view assumption holds in practice is dependent on the choice of views and the downstream tasks, and its limitations are known~\citep{tsai2021selfsupervised,shwartzziv2023compress}.
Still, we use it as an idealized principle
that motivates a criterion for learning representations.

\emph{Setup:} Given a sample $X$, we apply masking, cropping, and photometric distortions to generate views.
For generality, we assume an asymmetric student-teacher setup: given paired views $X_S, X_T$, the teacher
defines a target
$T
\sim p(\cdot \mid X_T)$, and the student encoder $f$ outputs a representation $R$ for $X_S$.
To predict the target $T$ from $X_S$, a projection head maps the representation to a predictive distribution $q(\,\cdot\,\mid X_S)$ over the target space. In practice, $T$ is a \textit{collection of discrete variables} as discussed later.

\paragraph{InfoMax objective} The information we would like to capture in the representation $R$ is the student-teacher shared mutual information (MI) $I(X_S; X_T)$.
The teacher exposes information about its view $X_T$ through the target $T$.
By the data processing inequality, we have $I(X_S; X_T) \geq I(X_S; T) \geq I(R; T)$.
This yields the \emph{multi-view InfoMax} objective of maximizing $I(R; T)$, \ie, the shared information of the representation $R$ with the target $T$ (see \cref{fig:intro:info}).
We can maximize this objective using a variational lower bound~\citep{barber2003algorithm}, where $H(T)$ is the entropy of the target and $\KL$ is the Kullback-Leibler divergence (all derivations in \cref{app:sec:derivations}):
\begin{align}
    I(R; T) &\geq H(T) + \mathbb{E} [ \log q(T \mid X_S) ] \label{eqn:barber} \\
              &= \underbrace{I(X_T; T)}_\text{anti-collapse} - \underbrace{\mathbb{E}_{X_T,X_S} \KL\left( p(T \mid X_T) \mid\mid q(T \mid X_S) \right)}_\text{alignment}. \label{eqn:lower-bound}
\end{align}
The lower bound showcases two elements intuitively important for any SSL algorithm: an \emph{anti-collapse term}, measuring the amount of information the targets $T$ retain about the input $X_T$, and an \emph{alignment term}, measuring how well the student approximates the teacher.
In practice, we achieved the best performance with the teacher an EMA of the student, and therefore only optimize \cref{eqn:lower-bound} with respect to the student parameters.
Thus, the anti-collapse term $I(X_T; T)$ on the teacher side is not directly maximized, which we resolve in \cref{subsec:global-collapse-prevention,subsec:struct-collapse-prevention}.

\paragraph{Granularity of the objective} 
The KL term in \cref{eqn:lower-bound} is flexible regarding the granularity of the targets; 
for instance, we could align the two views through a single image-level target.
Instead, to leverage more information, we model the targets as a collection of $L$ elements $T=(T^{(1)}, \ldots, T^{(L)})$, where $\ell$ indexes the targets by patch and feature layer.
In this way, the training signal is both spatial and stems from features from different depths of the model.
In the following, we assume that student and teacher distributions factorize over the elements.
The alignment term in \cref{eqn:lower-bound} then decomposes into a sum of element-wise alignment terms.

\paragraph{Preventing collapse}
From \cref{eqn:lower-bound}, we can see that collapse occurs when $I(X_T; T)=H(T)-H(T \mid X_T)$ goes towards 0.
Preventing collapse thus involves maintaining targets with high \emph{marginal entropy} $H(T)$ and low \emph{conditional entropy} $H(T \mid X_T)$.
Intuitively, the targets should be spread around the data distribution, while remaining predictable for any particular input.
For collections of elements, $I(X_T; T)$ decomposes as
\begin{equation}
    I(X_T; T) = \sum_{\ell=1}^L \underbrace{I(X_T; T^{(\ell)})}_\text{per-element info} - \underbrace{\TC(T)}_\text{redundancy} + \underbrace{\TC(T \mid X_T)}_\text{0}, \label{eqn:anti-collapse-elementwise}
\end{equation}
where the \emph{total correlation} $\TC(T)$ measures the redundancy among elements and $TC(T \mid X_T)=0$ by assuming a factorized $p(T \mid X_T)$. Hence, preventing collapse can be achieved by maintaining high per-element information while limiting the uncontrolled growth of redundancy (\cref{subsec:global-collapse-prevention,subsec:struct-collapse-prevention}, respectively).

\subsection{\method: a New Joint-Embedding Multi-View Objective}
\begin{figure}[t!]
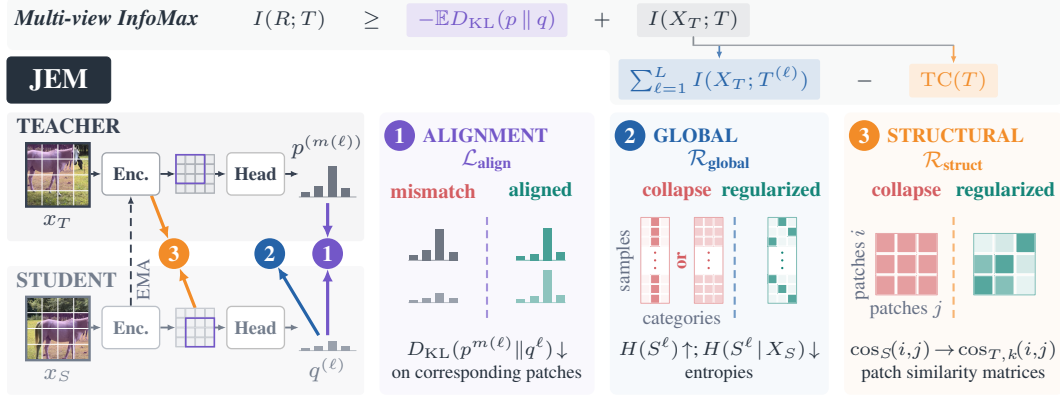

    \centering
    \includefigurepdf{figures/method.pdf}\space
    \vspace{-18pt}
    \caption{
    \textbf{From multi-view InfoMax to \method.}
    Maximizing the variational lower bound on the multi-view InfoMax objective in \cref{eqn:lower-bound} yields two requirements: cross-view alignment between teacher and student distributions $p$ and $q$, and informative, non-collapsed targets $T \sim q(\cdot \mid X_T)$.
    \method implements this with three loss terms:
    the KL term $\mathcal{L}_{\text{align}}$ aligns student patch $\ell$ with matching teacher target $m(\ell)$;
    $\mathcal R_{\text{global}}$ maintains per-patch mutual information through marginal and conditional entropy;
    and $\mathcal R_{\text{struct}}$ seeks to control the total correlation of the targets $\TC(T)$ by preserving the geometry of teacher features of layer $k$ in the final student representation. Please zoom for details.
    }
    \label{fig:method}
\end{figure}

In this section, we discuss our SSL method \method.
Motivated by maximizing the multi-view InfoMax objective, \method uses a principled loss function consisting of an alignment and regularization terms derived from \cref{eqn:lower-bound,eqn:anti-collapse-elementwise}.
We provide an overview of our method in \cref{fig:method}. 

\subsubsection{Patch-Wise Alignment Loss}
\label{subsec:alignemnt-loss}
For a given image, we produce a pair of overlapping student-teacher views $x_S$, $x_T$.
Both views are placed on the same patch grid so overlapping patches correspond one-to-one.
We denote $\mathcal O_S$ as the set of indices of student patch-layer pairs also appearing in the teacher view, and $m\colon \mathcal O_S \mapsto \mathcal O_T$ the function mapping the student indices to the corresponding teacher patch-layer pairs with indices $\mathcal O_T$.
This choice of mapping enforces an invariance to the index of the elements.
The alignment loss $\mathcal L_\text{align}$ is the element-wise decomposition of the KL in \cref{eqn:lower-bound} over $\mathcal O_S$:
\begin{equation}
    \mathcal L_\text{align}(x_S, x_T) = \frac{1}{|\mathcal O_S|}\sum_{\ell \in \mathcal O_S} D_\text{KL}\left( p(T^{(m(\ell))} \mid x_T) \mid\mid q(T^{(\ell)} \mid x_S) \right). \label{eqn:alignment-loss}
\end{equation}

\subsubsection{Preventing Global Collapse}
\label{subsec:global-collapse-prevention}

To maximize the lower-bound in \cref{eqn:lower-bound}, our goal is to keep $I(X_T; T)$ large. 
As we use discrete targets, directly estimating $I(X_T; T)$ is hard for multi-variate $T$.
Instead, we maximize element-wise information terms $I(X_T; T^{(\ell)})$, with $\ell$ the element index, using the decomposition in \cref{eqn:anti-collapse-elementwise}.
We implement this with an explicit regularization loss on the student. 
Through EMA updates, the student acts as a proxy for the teacher.
In particular, we maximize $I(X_S; S^{(\ell)})=H(S^{(\ell)}) - H(S^{(\ell)} \mid X_S)$, where $S^{(\ell)}$ follows the $K$-class categorical $q(\cdot \mid Z_S^{(\ell)})$ parametrized by the student head output $Z_S^{(\ell)}$ for patch $\ell$.
Given head outputs $\mathbf z^{(\ell)}$ for a batch of $N$ inputs $\mathbf x$, we can compute an estimate of the element-wise mutual information as:
\begin{equation}
    \hat{I}(\mathbf x; \mathbf z^{(\ell)}) = \underbrace{- \sum_{j=1}^K \bar{q}_j \log \bar{q}_j}_\text{marg. entropy} - (\underbrace{- \frac{1}{N} \sum_{i=1}^N \sum_{j=1}^K q_j(z_i^{(\ell)}) \log q_j(z_i^{(\ell)} )}_\text{cond. entropy}), \label{eqn:mutual-information-estimate}
\end{equation}
where $\bar{q}_j=\frac{1}{N}\sum_{i=1}^N q_j(z_i^{(\ell)})$ denotes the mean batch probability for class $j$ and patch $\ell$.
The global collapse regularization loss is then given as:
\begin{equation}
\mathcal R_\text{global}(\mathbf x) = -\frac{1}{L} \sum_{\ell=1}^{L} \hat{I}(\mathbf x; \mathbf z^{(\ell)}).
\label{eq:global_collapse_regularization}
\end{equation}
However, this regularization term only partially covers the anti-collapse term $I(X_T; T)$ in \cref{eqn:lower-bound} due to the element-wise computation of $I(X_S; S^{(\ell)})$.
We therefore define an additional regularization term which covers the \emph{structural} collapse mode.

\subsubsection{Preventing Structural Collapse}
\label{subsec:struct-collapse-prevention}
In the last section, we derived a regularizer to maintain the first term of \cref{eqn:anti-collapse-elementwise}, $\sum_{\ell=1}^L I(X_T; T^{(\ell)})$, preventing global collapse of the targets.
In this section, we address the second term, the redundancy $\TC(T)$.
Intuitively, the redundancy becomes large when all $T^{(\ell)}$ encode similar information, which we term a \emph{structural collapse}.
For example, if each patch target encodes the same input-dependent category, the jointly-encoded information $I(X_T; T)$ is the same as that of a single target $I(X_T; T^{(\ell)})$.
Maximizing the sum of element-wise information $I(X_T; T^{(\ell)})$ does not stop this collapse mode; thus, we design a mechanism to prevent the redundancy from growing too large.

Estimating the total correlation $\TC(T)$ of a large set of discrete variables is intractable.
Instead, we encourage the representation $R$ to retain the dependency structure contained in the input, using this as a proxy for \emph{controlling} $TC(T)$.
As the true dependencies are inaccessible, we use the similarities of representations from different layers of the teacher as a proxy; this rests on the assumption that different layers encode different levels of structure.
The representation must then match these similarities after a learned non-linear transformation.
Specifically, given student representation $r$, and teacher representation $h_k$ and transformation $g_k$ at layer $k$, the loss for a pair of elements at indices $i, j$ can be written as
\begin{equation}
\mathcal L_\text{struct}^{(i,j)}(r, h_k) = \left(\operatorname{cos}(g_k(r^{(i)}), g_k(r^{(j)})) - \operatorname{cos}(h_k^{(m(i))}, h_k^{(m(j))})\right)^2
\end{equation}
where $\operatorname{cos}$ denotes the cosine similarity and $(i,j)$ are restricted to the subset of elements coming from the same layer of $\mathcal O_S \times \mathcal O_S$.
Intuitively, by being able to reconstruct the early layer similarities, the representation avoids the collapse to a single, redundant vector, and is still free to encode higher-level semantic structure such as similar object patches.
Given a set of target teacher layers $\mathcal K$, the total loss is then averaged over target layers and patch pairs:
\begin{equation}
    \mathcal R_\text{struct}(x_S, x_T) = \frac{1}{|\mathcal K| |\mathcal O_S| (|\mathcal O_S|-1) } \sum_{k \in \mathcal K} \sum_{i \in \mathcal O_S} \sum_{\substack{j \in \mathcal O_S\\i\neq j}} \mathcal L_\text{struct}^{(i,j)}(r, h_k).
    \label{eq:structural_regularization}
\end{equation}

\subsection{View Construction}
\label{subsec:view}
While often relegated to implementation details, the algorithm used to construct views of the input is a critical design choice in SSL~\citep{chen2020simple,tian2020goodviews,he2021masked}.
For images, one obvious way to create views is cropping~\citep{bachman2019learning,chen2020simple}, commonly combined with photometric transformations~\citep{chen2020simple,grill2020bootstrap}.
The resulting crops overlap to varying degrees, leading to context-invariant representations. 
Additionally, \cite{caron2021swav, caron2021emerging} exploit smaller local crops, which induce local-to-global correspondences.
Masking-based methods, on the other hand, typically use fully overlapping student and teacher crops~\citep{he2021masked,assran2022masked}, where masking controls which patches are visible to the encoder. 

From the perspective of the multi-view assumption, \emph{cropping} and \emph{masking} are complementary mechanisms for controlling the information shared between views.
In this work, we treat them in a unified manner. 
We build several student views which partially overlap with the teacher view and are, in addition, masked, such that the views differ both in spatial support and patch visibility.
The alignment loss of \cref{eqn:alignment-loss} is applied to \emph{all overlapping patches}, whether masked or visible.
Note that traditional masked prediction is a special case of this construction, obtained with a single student view and the student/teacher crops coinciding.
Interestingly, we found that the method performs well without the use of masks, showing that patch-level SSL is possible using only cropping (see \cref{sec:exp:ablation}).

\begin{wrapfigure}{r}{0.32\textwidth}
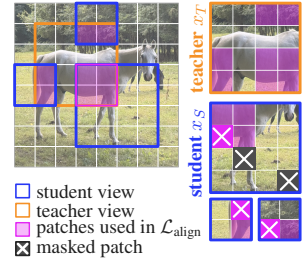

\vspace{-1.4\baselineskip}
\centering
\resizebox{0.95\linewidth}{!}{\includefigurepdf{figures/view_construction_inline.pdf}\space}
\captionsetup{font=small,skip=3pt}
\vspace{-0.5pt}
\caption{\textbf{View construction.}}
\label{fig:view-construction}
\vspace{-2\baselineskip}
\end{wrapfigure}

Concretely, we implement this design by sampling shifted but overlapping student and teacher crops aligned on a common patch grid, as shown in \cref{fig:view-construction}.
We integrate \emph{local crops}~\citep{caron2021swav} by sampling smaller student than teacher crops, obtaining local-to-global prediction pairs.  For a fraction of student views, we sample and apply a \emph{patch mask}.
Student and teacher views are also independently transformed with a standard set of photometric transformations~\citep{grill2020bootstrap}.  Further implementation details are provided in \cref{app:sec:view-construction}.

\section{Experiments}

This section presents the experimental results validating \method. We first provide our implementation details (\cref{sec:exp:impl-details}) and ablate the individual components (\cref{sec:exp:ablation}). We evaluate the quality of the resulting features against models from the literature (\cref{sec:exp:res}) and study scaling behaviour (\cref{sec:exp:scaling}).

\subsection{Implementation details}
\label{sec:exp:impl-details}

\paragraph{Architecture}
Our encoder is a Vision Transformer (ViT) \citep{dosovitskiy2021image}, at three model scales, ViT-L, ViT-g and ViT-7B.
All models use $16 \times 16$ patches, omit the class token, and prepend four learned register tokens \citep{darcet2024vision}.
We follow the open source implementation of \citet{simeoni2025dinov3}, and inherit most hyperparameters from the published DINOv3 configurations.

\paragraph{Training Details}
For each image, we sample two view sets, each containing a 
global student view, four local student views and one teacher view.
Global and teacher views are $256\times256$ pixels and local views
$112\times112$.
In $80\%$ of the global and local views we mask $65\%$ and $50\%$ of the patches respectively; the rest of the views are left unmasked.
We use $K=4096$ classes for the categorical distributions.
We train ViT-g for $500$k iterations with AdamW, a global batch size of $3072$, and a constant learning rate schedule with linear warm-up.
The teacher is an exponential moving average of the student with fixed momentum $0.994$.
The global collapse term $\mathcal R_\text{global}$ has a coefficient of $0.1$ with a linear decay from $1$ over the first $12.5$k steps.
The structural collapse term $\mathcal R_\text{struct}$ has a coefficient increased linearly from $0.05$ to $1$ over the course of training, with a linear warm-up from 0 over the first $12.5$k steps.
More hyperparameter details 
in \cref{app:sec:training-details}.

We train on the public ImageNet-1k \citep{russakovsky2015imagenet} and ImageNet-22k \citep{deng2009imagenet} datasets for comparability with prior work. 
For larger scale, we follow \citet{fu2026lingbotvision} and reproduce the curated LVD-142M dataset \citep{oquab2024dinov2}, denoted ``\dataset'' (\cref{app:sec:details-data} for details).

\begin{table*}[t]
    \centering
    \small
    \sisetup{detect-all, detect-weight=true, detect-inline-weight=math, table-number-alignment=center}
    \renewcommand{\arraystretch}{1.08}
    \captionsetup[subtable]{
        position=top,
        justification=centering,
        singlelinecheck=true,
        font=small,
        labelfont=bf,
        skip=4pt
    }

    \caption{\textbf{Ablation study.}
    We isolate the contributions of view construction (a),
    the composition of the patch-level KL alignment loss (b),
    and the terms in the complete objective (c).  We train ViT-L models on ImageNet-22k for 250k steps and report attention-probe accuracy on IN1k and segmentation mIoU on ADE20k.
    Tables compare the full model with variants removing (w/o) individual components.
    }
    \label{tab:component-ablations}
    \vspace{-6pt}

    \begin{minipage}[t]{0.33\textwidth}
        \centering
        \subcaption{View construction.}
        \label{tab:ablation-view-construction}
        \setlength{\tabcolsep}{3pt}
        \begin{tabular}{@{}l@{\hspace{6pt}}S[table-format=2.1]S[table-format=2.1]@{}}
            \toprule
            Configuration & {IN1k} & {ADE} \\
            \midrule
            \textit{Full \method model}            & \bfseries 82.8 & \bfseries 48.1 \\
            w/o~local crops                  & 81.2 & 46.2 \\
            w/o~masking                     & 82.3 & 45.3 \\
            w/o~stud.-teach. shift       & 82.8 & 46.6 \\
            w/o~independ. aug.              & 74.8 & 31.4 \\
            \bottomrule
        \end{tabular}
    \end{minipage}
    \hfill
    \begin{minipage}[t]{0.33\textwidth}
        \centering
        \subcaption{KL alignment.}
        \label{tab:ablation-kl-composition}
        \setlength{\tabcolsep}{3pt}
        \begin{tabular}{@{}l@{\hspace{6pt}}S[table-format=2.1]S[table-format=2.1]@{}}
            \toprule
            Configuration & {IN1k} & {ADE} \\
            \midrule
            \textit{Full \method model}           & 82.8 &  48.1 \\
            w/o~multi-layer loss          & 81.7 & 48.3 \\
            w/o~visible-patch loss        & 70.8 & 28.1 \\
            w/o~masked-patch loss         & 81.3 & 46.0 \\
            w/o~EMA       & 83.0 & 47.0 \\
            \bottomrule
        \end{tabular}
    \end{minipage}
    \hfill
    \begin{minipage}[t]{0.31\textwidth}
        \centering
        \subcaption{Objective.}
        \label{tab:ablation-loss-components}
        \setlength{\tabcolsep}{3.8pt}
        \begin{tabular}{@{}l@{\hspace{6pt}}S[table-format=2.1]S[table-format=2.1]@{}}
            \toprule
            Configuration & {IN1k} & {ADE} \\
            \midrule
            \textit{Full \method model}          & \bfseries 82.8 & \bfseries 48.1 \\
            w/o~$\mathcal{L}_{\text{align}}$                 &  63.1 & 22.4 \\
            w/o~$\mathcal{R}_{\text{global}}$
            & 10.3 & 0.5 \\
            w/o~$\mathcal{R}_{\text{struct}}$
            & 82.3 & 48.0 \\
            \bottomrule
        \end{tabular}
    \end{minipage}
\end{table*}

\vspace{-0.5em plus 0.5em}
\paragraph{Evaluation Protocol}
We evaluate all methods with a frozen backbone and simple probing setups.
For classification, we report attention probe accuracy on ImageNet-1k \citep{russakovsky2015imagenet}.
For semantic segmentation, we report linear probe mIoU on ADE20k \citep{zhou2019ade20k}, Pascal VOC \citep{everingham2010pascal}, and Cityscapes \citep{cordts2016cityscapes}; for monocular depth, we report linear probe RMSE on NYUv2 \citep{silberman2012nyuv2} and KITTI \citep{geiger2012kitti}.
We also consider more challenging benchmarks testing combined class- and instance understanding, and report mask AP for instance segmentation on COCO~\citep{lin2014coco}, and PQ for panoptic segmentation on COCONut~\citep{deng2024coconut} and ADE20k.
See \cref{app:sec:benchmark-protocols} for details.

\subsection{Ablations}
\label{sec:exp:ablation}

In this section, we study the impact of the different components of \method. 
In all experiments, if not stated otherwise, we train ViT-L-sized models on the ImageNet-22k dataset for 250k iterations. 
We evaluate on ImageNet-1k (IN1k) classification and ADE20k (ADE) semantic segmentation.

\paragraph[View construction]{View construction (\cref{tab:ablation-view-construction})}
Overall, we find \method to be robustness to individual changes in the view construction: \method can deliver strong performance without using masked patches, local crops, and shifting student and teacher global views (\ie fully overlapping views), vital components of other methods~\citep{caron2021emerging,oquab2024dinov2}.
Still, these techniques are complementary and all needed to achieve good overall performance.
However, removing per-view independent photometric augmentations such as random color shifts does impact \method and greatly degrades the performance on both global and dense tasks.
We believe this is because these augmentations induce an invariance to pixel-level details, enabling the model to learn semantic local features instead of a shortcut solution.

\paragraph[Alignment loss]{Alignment loss (\cref{tab:ablation-kl-composition})} 
Multi-layer prediction, masked-patch prediction, and an EMA teacher are not individually essential for strong performance: removing any one of them changes IN1k accuracy by at most 1.5 points and ADE20k mIoU by at most 2.1 points.
This is remarkable, as masked prediction is usually seen as a key component of patch-level SSL.
Similarly, EMA teachers were previously critical to enable both self-clustering~\citep{caron2021emerging} and latent-space masked prediction~\citep{assran2023self}.
However, it is important to include the visible patches in $\mathcal{L}_\text{align}$; removing them costs 20 points on ADE20k.
We hypothesize this is because the loss prevents the patch representation from drifting away from spatial alignment.

\begin{figure*}[t]
    \centering
    \begin{minipage}[t]{0.35\textwidth}
        \centering
        \includefigurepdf{figures/entropy-weight-ablation.pdf}\space\space
        \vspace{-5pt}
        \caption{\textbf{Impact of the global regularization weight} on IN-1k and ADE20k performance. Models use ViT-L, IN-22k, and 250k steps.}
        \label{fig:reg:loss-teacher-entropy}
    \end{minipage}
    \hfill
    \begin{minipage}[t]{0.63\textwidth}
        \centering
        \includefigurepdf{figures/loss_terms_teacher_entropy.pdf}\space
        \vspace{-5pt}
        \caption{\textbf{Collapse prevention.} (Left) Teacher patch mutual information (MI) collapses when disabling $\mathcal{R}_{\text{global}}$. (Right) Mean $\ell_2$-distance between patches and the image mean patch representation; low means loss of spatial structure. 
        ViT-g on A140M.}
        \label{fig:reg:collapse-prevention}
    \end{minipage}
\end{figure*}

\paragraph[Objective]{Objective (\cref{tab:ablation-loss-components})}
We ablate the three terms of our objective.
First, removing the \emph{alignment loss} $\mathcal{L}_{\text{align}}$~(\cref{eqn:alignment-loss}) sharply reduces IN1k accuracy from 82.8 to 63.1 and ADE20k mIoU from 48.1 to 22.4, showing that the alignment loss carries a critical training signal. 
Second, removing \emph{global collapse regularization} $\mathcal{R}_{\text{global}}$ (\cref{eq:global_collapse_regularization}) is catastrophic, and leads the model to collapse, attaining 10.3 IN1k accuracy.
In \cref{fig:reg:loss-teacher-entropy}, we also examine how sensitive performance is to this regularizer's loss weight.
We observe that the concrete weight matters only in that it must be non-zero: a weight of $0$ yields degenerate representations, while any value across a wide range performs comparably. Third, we find the effect of \emph{structural regularization}~$\mathcal{R}_{\text{struct}}$ (\cref{eq:structural_regularization}) to be marginal at ViT-L scale (-0.5 on IN1k), but crucial when scaling the model size further.
Next, we investigate its effect with larger model sizes, and find that it is important to prevent the progressive loss of spatial diversity.

\paragraph[Preventing collapse]{Preventing collapse (\cref{fig:reg:collapse-prevention})} 
Now we consider larger models (ViT-g) which suffer more and faster from collapse.
First, in \cref{fig:reg:collapse-prevention}~(left), we show the evolution of the per-patch MI $I(X_T; T^{(\ell)})$ over training when disabling global regularization $\mathcal{R}_{\text{global}}$; in this case, the target distributions collapse to a uniform or constant distribution over categories.
Second, in \cref{fig:reg:collapse-prevention}~(right), we show how the average patch $\ell_2$-distance to the per-image mean patch develops over training with or without the structural regularizer. 
Low scores indicate that patch representations become more spatially uniform. 
Without regularization, increasing degradation starts after 250k iterations, leading to performance degradation.
In contrast, the regularized model keeps patch representations distinct from their spatial average, thereby maintaining stronger performance on dense tasks (+2.7 ADE20k mIou).

\begin{table*}[t]
    \centering
    \small
    \setlength{\tabcolsep}{1.75pt}
    \sisetup{detect-all, detect-weight=true, detect-inline-weight=math, table-number-alignment=center}

    \caption{\textbf{Comparison to prior SSL methods}.
    Classification reports IN-1k attention-probe accuracy.
    Semantic, instance, and panoptic segmentation results are mIoU, mAP, and PQ, respectively.
    Depth reports RMSE (lower better).
    Tokens is the approximate number of tokens seen during training in trillions.
    `*' denotes the DINOv2 algorithm trained with the \method setup for 125k steps on IN-1k and 500k steps on IN-22k or \dataset.
    $\dagger$ denotes backbone training or fine-tuning above 256 px. 
    }
    \vspace{-5pt}
    \label{tab:in-results}
    \fontsize{8.75}{10}\selectfont
    \setlength{\tabcolsep}{1.25pt}
    
    \begin{tabular}{@{}lll@{\hspace{0.5pt}}
        S[table-format=1.2]
        S[table-format=2.1]
        S[table-format=2.1]
        S[table-format=2.1]
        S[table-format=2.1]
        S[table-format=2.1]
        S[table-format=2.1]
        S[table-format=2.1]
        S[table-format=1.3]
        S[table-format=1.3]@{}}
    \toprule
 & & & & \multicolumn{1}{c}{Class.} & \multicolumn{3}{c}{Sem. Seg.} & \multicolumn{1}{c}{Inst. Seg.} & \multicolumn{2}{c}{Pan. Seg.} & \multicolumn{2}{c}{Depth} \\
    \cmidrule(lr){5-5} \cmidrule(lr){6-8} \cmidrule(lr){9-9} \cmidrule(lr){10-11} \cmidrule(l){12-13}
    Model & Size & Dataset & {Tokens (T)} & {IN1k} & {ADE} & {VOC} & {City.} & {COCO} & {ADE} & {COCONut} & {NYU} & {KITTI} \\
    \midrule
 ~~MAE & L & IN-1k & 0.40 & 78.4 & 31.6 & 66.1 & 53.1 & 6.0 & 6.5 & 9.7 & 0.489 & 2.903 \\ ~~MSN & L & IN-1k & 0.43 & 76.3 & 24.3 & 55.0 & 46.4 & 3.2 & 4.4 & 6.7 & 0.656 & 3.696 \\ ~~data2vec 2.0 & L & IN-1k & 0.60 & 79.5 & 27.4 & 52.7 & 48.6 & 2.1 & 4.8 & 7.1 & 0.512 & 2.901 \\ ~~CAPI & L & IN-1k & 2.10 & 82.9 & 31.9 & 66.0 & 50.6 & 10.7 & 12.7 & 14.7 & \bfseries 0.361 & \bfseries 2.604 \\ ~~DINOv2* & L & IN-1k & 0.23 & \bfseries 83.8 & 45.7 & 83.3 & 66.5 & 12.1 & 18.0 & 22.1 & 0.408 & 2.718 \\ ~~\textbf{\methodNormal} & L & IN-1k & 0.23 & 82.8 & \bfseries 46.4 & \bfseries 84.9 & \bfseries 67.4 & \bfseries 13.5 & \bfseries 20.1 & \bfseries 24.3 & 0.396 & 2.627 \\

    \midrule
 ~~I-JEPA & g & IN-22k & 0.12 & 78.8 & 34.2 & 73.6 & 52.4 & 6.3 & 7.4 & 10.0 & 0.471 & 2.994 \\ ~~Franca & g & IN-22k & 1.74 & 86.0 & 45.9 & 81.9 & 67.0 & 11.9 & 13.9 & 16.4 & 0.355 & 2.505 \\ ~~DINOv2* & g & IN-22k & 1.39 & \bfseries 86.5 & 50.0 & 83.6 & 68.5 & 13.0 & 18.1 & 21.1 & \bfseries 0.343 & 2.494 \\ ~~\textbf{\methodNormal} & g & IN-22k & 1.39 & 85.5 & \bfseries 51.2 & \bfseries 85.9 & \bfseries 69.6 & \bfseries 14.7 & \bfseries 22.4 & \bfseries 25.7 & 0.363 & \bfseries 2.472 \\ 

    \midrule
 \multicolumn{13}{@{}l}{\textit{$>$1B, trained on large datasets}} \\
 ~~DINOv2* & g & \dataset & 1.39 & 86.9 & 50.5 & 83.6 & 69.4 & 12.7 & 18.7 & 21.4 & 0.339 & 2.455 \\ ~~DINOv2 (reg.)$^\dagger$ & g & LVD-142M & 1.83 & 87.3 & 49.5 & 83.1 & 69.1 & 12.3 & 16.7 & 18.6 & 0.357 & 2.569 \\ ~~LingBot-Vision$^\dagger$ & g & Unk.-163M & 1.86 & 86.5 & 53.6 & 85.9 & 74.1 & 15.2 & 22.2 & 25.0 & 0.287 & 2.127 \\ ~~\textbf{\methodNormal} & g & \dataset & 1.39 & 85.6 & 52.3 & 86.1 & 71.2 & 14.7 & 22.9 & 26.1 & 0.329 & 2.324 \\ ~~V-JEPA 2.1$^\dagger$ & G & VM-163M & 0.24 & 83.2 & 47.8 & 84.3 & 68.3 & 12.5 & 15.5 & 19.6 & \bfseries 0.277 & 2.167 \\ ~~WebSSL & 7B & MC-8B & 7.23 & 87.5 & 42.2 & 75.7 & 61.5 & 9.3 & 9.9 & 13.7 & 0.460 & 2.945 \\ ~~DINOv2* & 7B & \dataset & 1.85 & 87.7 & 51.9 & 84.5 & 71.1 & 14.0 & 17.9 & 22.0 & 0.289 & 2.218 \\ ~~DINOv3$^\dagger$ & 7B & LVD-1689M & 4.06 & \bfseries 89.0 & \bfseries 55.9 & 86.7 & \bfseries 75.2 & \bfseries 16.0 & 21.6 & 27.0 & 0.278 & 2.242 \\ ~~\textbf{\methodNormal} & 7B & \dataset & 1.85 & 86.9 & 54.4 & \bfseries 87.1 & 73.6 & 15.6 & \bfseries 23.9 & \bfseries 27.3 & 0.297 & \bfseries 2.125 \\ 

    \bottomrule
    \end{tabular}
    \vspace{-5pt}

\end{table*}

\subsection{Comparison to Other SSL Algorithms}
\label{sec:exp:res}
In \cref{tab:in-results}, we compare \method to other SSL methods on the same data, training ViT-L on ImageNet-1k and ViT-g on ImageNet-22k. 
We also report results with models trained on larger, not directly comparable datasets and discuss them in \cref{sec:exp:scaling}. 
For a controlled comparison, we re-train DINOv2 models~\citep{oquab2024dinov2} (denoted ``DINOv2*''), using the same setup as \method (similar to training DINOv3 without gram and high-resolution finetuning steps). 
This baseline is relevant as DINOv2, combining global and local losses, serves as a foundation for many of today's best SSL methods.

For both model sizes, \method outperforms all other methods on six out of nine tasks, with consistent gains across the full range of dense prediction: from coarse semantic segmentation to fine-grained instance and panoptic segmentation. 
On global classification, we observe a small gap to DINOv2; still a strong result as DINOv2 was heavily designed to obtain the best global performance.
Interestingly, \method scales effortlessly to ViT-g. 
Moreover, its advantage over DINOv2 increases at scale, \eg with a 4 PQ pt gain on both challenging panoptic segmentation tasks; the gap is maintained at 7B (\cref{sec:exp:scaling}).

\subsection{Scaling \method}
\label{sec:exp:scaling}

An important aspect of modern ML algorithms is their scalability with model size and data.
In SSL, scaling is often challenging due to feedback dynamics of self-supervision; different model scales show different behavior~\citep{simeoni2025dinov3}, and approaches that work well on small scales do not necessarily train well (or at all) beyond 1B parameters. Here we show that \method scales well. 

\begin{figure}[!t]
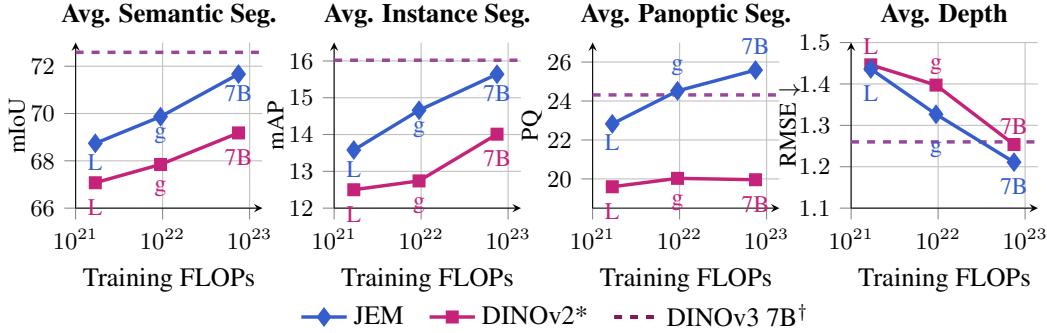

    \centering
    \includefigurepdf{figures/scaling-flops-averages.pdf}\space
    \vspace{-25pt}
    \caption{\textbf{Scaling model size.}   
    Performance as a function of approximate training compute, averaged over datasets for each benchmark type. DINOv2* and \method are trained by us for 500k steps on \dataset. 
    The dashed purple lines show DINOv3 ViT-7B.
    }
    \label{fig:scaling-flops-averages}
    \vspace{-8pt}
\end{figure}

We train 300M (ViT-L), 1B (ViT-g), and 7B (ViT-7B) parameter models with \method and DINOv2 on our \dataset dataset for 500k steps.
\cref{fig:scaling-flops-averages} plots task-averaged performance as a function of training FLOPs (individual benchmarks in appendix \cref{fig:scaling-flops}). 
We also include DINOv3 ViT-7B results (dashed line), though we note it is trained for twice as long on a 12$\times$ larger dataset and with additional refinement steps (gram and high-resolution fine-tuning). 
\method trains stably even at 7B scale, with performance improving significantly when model size grows.
To our knowledge, this is the first exclusively patch-level latent-space method that trains well at this scale.
Compared to the DINOv2 baseline, \method exhibits comparable or better scaling behavior and outperforms it at all model scales. 

\cref{tab:in-results} also lists models above 1B parameters trained on different large datasets which are thus not directly comparable. 
Notably, \method outperforms the best SSL method DINOv3 on several datasets in semantic and panoptic segmentation, as well as on depth estimation, despite DINOv3's longer training, larger dataset and its refinement steps (gram anchoring and high resolution fine-tuning)---showing the strength of our training paradigm. 
In \cref{fig:pca-comparison,fig:cosine-comparison}, we also plot qualitative PCA and patch cosine similarity results highlighting the quality of the patch features, obtained without any refinement steps.

\section{Conclusion}

In this work, we introduced \method, a principled and stable patch-level joint-embedding multi-view method that learns strong representations.
\method is derived from a multi-view InfoMax objective motivated by the multi-view assumption. This objective leads to three loss terms: one aligning representations across views, one preventing global collapse and one preserving structural integrity. \method trains stably from 300M to 7B parameters---to our knowledge, the first exclusively patch-level latent-space method demonstrated at this scale---and achieves impressive results on segmentation tasks while being competitive on classification and depth estimation.
These results suggest that multi-objective SSL's complexity is not necessary for strong performance, and establish principled patch-level objectives as a promising foundation for scalable visual representation learning.

\appendix
\clearpage
\section{Appendix}
\captionsetup[table]{position=top}
\captionsetup[subtable]{position=top,skip=4pt}

\noindent In this appendix, we provide additional analyses, implementation details, and experiments for \method{}.

\noindent Specifically:
\begin{itemize}[leftmargin=1.5em]
    \item \textbf{Additional discussion}. In \cref{app:sec:extended-related-work}, we discuss how existing self-supervised methods can be interpreted from the perspective of the lower bound of the multi-view InfoMax objective, and relate \method to more prior work.
    \item \textbf{Additional results}. In \cref{app:sec:additional:results}, we report scaling trends on individual benchmarks.
    \item \textbf{Additional details}. In \cref{app:sec:additional-details}, we describe the structural collapse regularizer, view construction, training procedure, and training data.
    \item \textbf{Derivations}. In \cref{app:sec:derivations}, we provide the full derivations of our objective.
    \item \textbf{Evaluation details}. In \cref{app:sec:evaluation-details}, we describe the evaluation protocol of each benchmark, including the implementation and reproduction of competing methods, list the evaluated model checkpoints, and detail the estimated number of training tokens for baselines.
    \item \textbf{Training configurations}. In \cref{app:sec:training-configurations}, we collect the complete architecture, training hyperparameter, and ablation settings tables.
    \item \textbf{Qualitative examples}. In \cref{app:sec:qualitative-examples}, we report additional qualitative feature visualizations of \method{} and competing models.
\end{itemize}

\subsection{Relationship to existing SSL paradigms}
\label{app:sec:extended-related-work}

\noindent In this section, we discuss how existing self-supervised methods can be interpreted from the multi-view InfoMax perspective. Further, we relate \method{} to prior works in which some of its key components also appear, namely: explicit entropy regularization of discrete targets, dense losses between partially overlapping views, multi-layer targets, and structural regularization, and discuss how our formulation differs. 

\paragraph{How different methods fit the multi-view InfoMax objective}
Self-clustering methods such as SwAV~\citep{caron2021swav}, DINO~\citep{caron2021emerging}, or iBOT~\citep{zhou2021ibot} use discrete categorical targets $T$, minimizing a cross entropy loss between student and teacher views that matches the KL divergence in \cref{eqn:lower-bound} up to a constant.
For the anti-collapse term, centering or Sinkhorn-Knopp balancing of the targets plays the role of maintaining $H(T)$; temperature sharpening makes sure that $H(T \mid X_T)$ does not grow too large.
In SimDINO~\citep{wu2025simplifying}, the targets are continuous vectors, with the minimized mean-squared-error (MSE) loss again matching the KL divergence; $H(T)$ is maintained by maximizing the log-determinant of the covariance matrix of the head outputs, and $H(T \mid X_T=x)$ is minimal for all $x$.
Similarly, redundancy-reduction methods~\citep{zbontar2021barlowtwins,bardes2021vicreg,balestriero2025lejepa} 
combine cross-view alignment with output-distribution regularization that can be seen as an indirect proxy for maintaining $H(T)$.

\paragraph{A principled view of the DINO family of methods.}
Owing to its strong performance, the DINO family of self-distillation methods~\citep{caron2021emerging,zhou2021ibot,oquab2024dinov2,simeoni2025dinov3} has become a common foundation for many recent methods in the literature~\citep{wu2025simplifying,fan2025webssl,venkataramanan2026franca}.
As discussed above, these methods can be interpreted as instantiations of the multi-view InfoMax objective, in which the cross-entropy between student and teacher implements the alignment term, while the anti-collapse term is only maintained implicitly, through mechanisms that were introduced empirically~\citep{caron2021swav,caron2021emerging}. In \method{}, each component of the training objective instead follows from the lower bound in \cref{eqn:lower-bound}, which simplifies the recipe of DINOv2 in three ways.
(i)~\emph{A single loss and single head for the final layer.} DINOv2 applies the DINO loss to the class token and the iBOT loss to the masked patches, with separate heads, and adds the KoLeo regularizer~\citep{sablayrolles2018spreading}, a nearest-neighbor estimator of the differential entropy of the class-token features, which acts as a further proxy for $H(T)$. \method{} instead applies the same alignment and anti-collapse terms to all elements of $T$, through one head at each supervised depth.
(ii)~\emph{Explicit regularization instead of heuristics.} Sinkhorn-Knopp balancing/centering and temperature sharpening indirectly control the marginal entropy $H(T)$ and the conditional entropy $H(T \mid X_T)$, respectively. \method{} replaces them with $\mathcal R_\text{global}$, which optimizes both quantities explicitly, so that student and teacher can share the same constant temperature, without a teacher temperature schedule.
(iii)~\emph{No prototype freezing.} DINO and DINOv2 freeze the last layer of the head, which holds the prototypes, during the first iterations of training to stabilize it; \method{} requires no such mechanism.
Moreover, the class token fits naturally into our information-theoretic derivation as \emph{one more element} of $T$, to which the same alignment and anti-collapse terms apply. From this perspective, DINO corresponds to the special case in which the class token is the only element of $T$, and iBOT and DINOv2 to adding the masked patches as further elements, each with its own loss and head.

\paragraph{Explicit entropy regularization of discrete assignments.}
For discrete targets, the InfoMax anti-collapse term $I(X_T; T) = H(T) - H(T \mid X_T)$ corresponds to maximizing the entropy of the average prediction while minimizing the entropy of each individual prediction.
Early works~\citep{bridle1991unsupervised,gomes2010discriminative} employ explicit mutual information maximization for training unsupervised classifiers.
More recently, unsupervised clustering methods have employed similar regularization techniques. SCAN~\citep{vangansbeke2020scan} maximizes the entropy of the mean prediction to avoid collapse of the cluster assignments, without an explicit term for the sharpness of individual predictions, which is only encouraged implicitly by its consistency loss between neighboring samples.
IIC~\citep{ji2019invariant} instead directly maximizes the mutual information between the cluster assignments of two views, without explicit regularization for each input's assignment.
IMSAT~\citep{hu2017imsat} includes both marginal and conditional entropy terms, combined with an alignment loss to enforce augmentation invariance.
Among self-clustering methods, MSN~\citep{assran2022masked} optimizes the marginal entropy, but also applies Sinkhorn-Knopp balancing to the targets and resorts to temperature sharpening to control the conditional entropy.

More similar to ours, TWIST~\citep{wang2021twist} employs an alignment loss between teacher and student assignment distributions, alongside the entropy regularization terms.
It is the closest image-level analogue of our $\mathcal R_\text{global}$ combined with $\mathcal L_\text{align}$.
In contrast to these works, our $\mathcal R_\text{global}$ is applied to the discrete assignments of every patch in a patch-level student-teacher framework.
Moving from per-image to per-patch assignment also introduces two collapse modes that entropy terms pooled over all patches cannot control: \textit{structural collapse} and \textit{positional collapse}.
The latter occurs when the model simply encodes the position of each patch in the grid. 
The marginal entropy over \emph{all} patches can be increased simply by encoding the patch position.
Instead, computing the marginal entropy separately for each patch position leads to estimating $I(X_T; T \mid \text{Pos})$, which prevents this failure mode.
In our derivation, this formulation naturally follows the decomposition in \cref{eqn:anti-collapse-elementwise}.
Among patch-level methods, CAPI~\citep{darcet2025capi} identifies the same positional collapse and alleviates it implicitly, by running the Sinkhorn-Knopp balancing of the targets separately at each patch position.
On the other hand, the \textit{structural collapse} failure mode entails patches within an image falling to an input-dependent constant distribution, effectively encoding the same image-level content.
This motivates our structural regularizer $\mathcal R_\text{struct}$ (\cref{subsec:struct-collapse-prevention}). 
Both of these terms have no counterpart in the image-level methods mentioned above.

\paragraph{Dense losses between partially overlapping views.}
A large family of dense SSL methods is based on masked image modeling, which predicts the content of masked patches, either in pixel~\citep{he2021masked}, or latent space~\citep{baevski2022data2vec,assran2023self}.
In pure masking methods, input and target fully overlap, whereas in our formulation we treat masking and cropping as complementary strategies to control information sharing between views. Therefore,
we utilize \textit{partially overlapping} views for teacher and student, additionally masking the student views.
Another category of methods applies joint-embedding losses between corresponding locations of two partially overlapping crops, but without masking.
Among these, PixPro~\citep{xie2021propagate} and VADeR~\citep{pinheiro2020unsupervised} are pixel-level contrastive methods.
Several patch-level clustering-based methods also rely on partially overlapping crops, aligning the patch assignments of the overlapping region with RoIAlign~\citep{ziegler2022selfsupervised,wen2022selfsupervised} or clustering the patches of both views jointly~\citep{stegmuller2023croc}.
VICRegL~\citep{bardes2022vicregl} applies a similar idea to continuous embeddings, matching each feature to its spatial nearest neighbor in the other view.
A few methods combine the two families, applying masked modeling across partially overlapping views, like ours.
LOCA~\citep{caron2024location} trains a student on masked query crops to predict teacher cluster assignments for a larger reference crop, where correspondences are interpolated.
SIM~\citep{tao2023siamese} uses a decoder to predict, from the visible patches of a masked view, the features of \emph{all} patches of a different view.
Differently from all these prior works, we sample student and teacher crops at integer patch offsets on a shared patch lattice, so that overlapping patches correspond exactly one-to-one; we found that interpolating targets between misaligned patch grids degrades the resulting dense features.
Moreover, we treat masked and visible patches uniformly, applying the alignment loss to all overlapping patches. 

\paragraph{Multiple layers as targets.}
Supervising intermediate layers dates back to deeply-supervised networks~\citep{lee2015deeply}, and to intermediate layer distillation~\citep{romero2015fitnets}.
In SSL, data2vec~\citep{baevski2022data2vec,baevski2023data2vec2} utilizes projections to regress the top-$K$ layers of an EMA teacher from the final layer of the student.
Among masking models, DeepMIM~\citep{ren2023deepmim} and LocalMIM~\citep{wang2023masked} add masked prediction losses on intermediate encoder layers, using as targets either pixels, image gradients, or teacher features.
More recently, V-JEPA~2.1~\citep{murlabadia2026vjepa21} fuses together the features of several student layers, for a shared predictor to regress the EMA-teacher features at the different layers, for both masked and visible tokens.
Similarly, Bootleg~\citep{lowe2026bootleg} extends I-JEPA by regressing, from the final student layer through a single predictor, the concatenated EMA-teacher features at the masked locations from different layers.
In \method{}, targets are indexed by patch \emph{and} layer: each supervised student layer $k$ predicts the categorical distribution computed by the EMA teacher at the \emph{same} depth $k$, so that the prediction at depth $k$ only depends on the first $k$ layers, and the alignment and anti-collapse terms are applied at every depth.
In addition, $\mathcal R_\text{struct}$ uses intermediate teacher layers as targets for the \emph{final} student layer, but only through their similarity structure (see next paragraph).

\paragraph{Structural regularization and Gram anchoring.}
Our structural regularizer $\mathcal R_\text{struct}$ (\cref{subsec:struct-collapse-prevention}) addresses the redundancy term $\TC(T)$ of the anti-collapse objective, whose uncontrolled growth leads to \emph{structural collapse}, in which all patches of an image encode the same image-level content.
We implement this regularizer by matching the pairwise cosine similarities between patches of a set of early-to-middle teacher layers from a projection of the features from the final student layer.
Using pairwise similarities as a target has also been employed by other methods.
In SSL, SERE~\citep{li2023sere} extends iBOT with a patch self-relation loss, training the similarities between the final-layer patch features of the student to match those of the EMA teacher on another view.
More recently, DINOv3~\citep{simeoni2025dinov3} adopts gram anchoring, to address the different phenomenon of the local patch structure becoming increasingly noisy with long training.
In DINOv3, after $1M$ iterations, an earlier checkpoint is used as a \emph{gram teacher}, to anchor the similarity matrix of the final-layer patch features of the student.
Differently, our targets come from \emph{early-to-middle} teacher layers and are only matched after a learned projection, so the final representation is not \emph{forced} to share the geometry of the earlier layers, but must retain the information required to recover it.
Moreover, our $\mathcal R_\text{struct}$ is applied from the start of training, uniformly on visible and masked patches, as well as local crops.

\subsection{Additional Results}
\label{app:sec:additional:results}

\begin{figure*}[t]
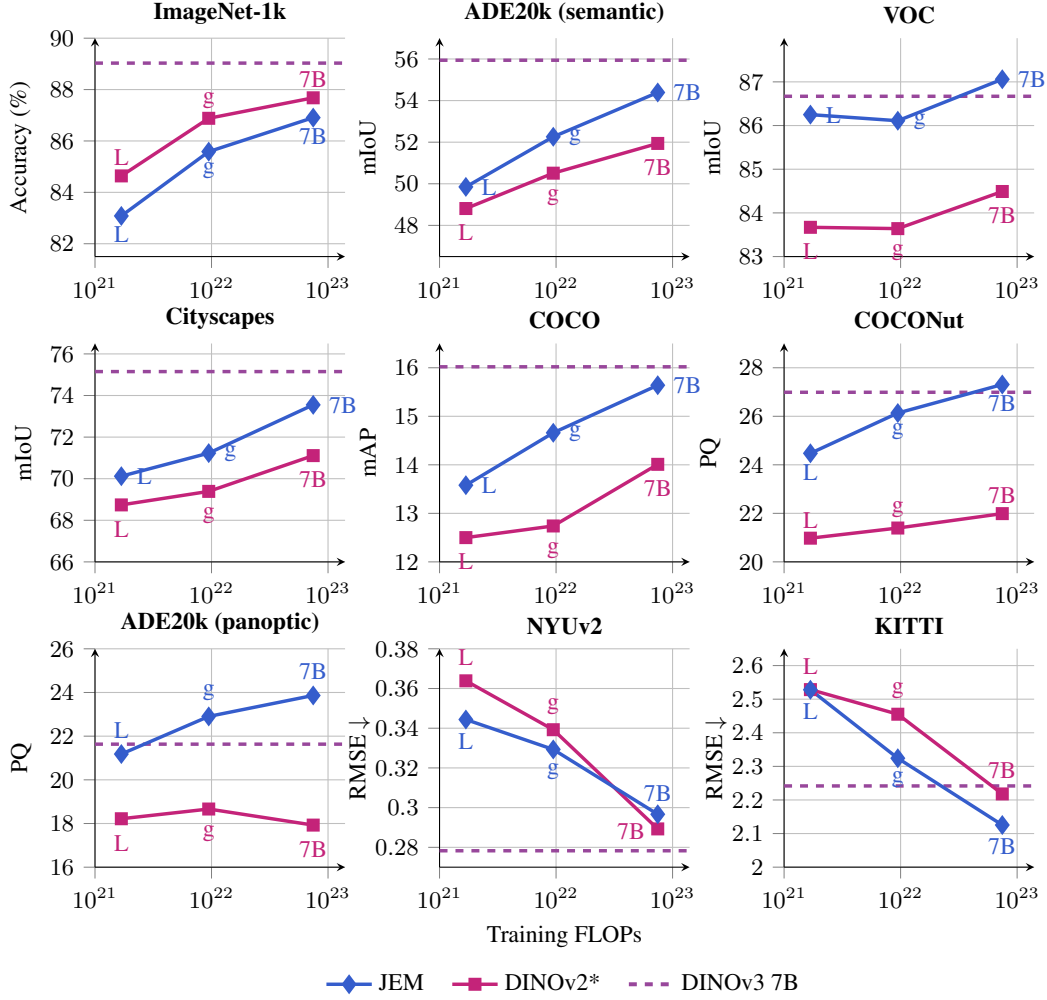

    \centering
    \includefigurepdf{figures/scaling-flops.pdf}\space
    \vspace{-12pt}
    \caption{\textbf{Scaling across individual benchmarks.} Performance as a function of approximate training compute. DINOv2* and \method are trained by us on \dataset for 500k steps. The purple dashed lines show DINOv3 ViT-7B. ADE20k is evaluated with both semantic segmentation (mIoU) and panoptic segmentation (PQ) probes.}
    \label{fig:scaling-flops}
\end{figure*}

\cref{fig:scaling-flops} reports the individual benchmark results underlying the task averages in \cref{fig:scaling-flops-averages}, together with ImageNet-1k classification accuracy.

\subsection{Implementation Details}
\label{app:sec:additional-details}

In this section, we give additional details about the structural collapse regularizer (\cref{app:sec:structural-collapse-regularizer}), construction of the views (\cref{app:sec:view-construction}), hyperparameters and training details (\cref{app:sec:training-details}), and the used training data (\cref{app:sec:details-data}).

\subsubsection{Structural Collapse Regularizer}
\label{app:sec:structural-collapse-regularizer}

We detail the structural collapse regularizer introduced in \cref{subsec:struct-collapse-prevention}.
For student and teacher views $x_S, x_T$, we define $\mathcal O_S$ and $\mathcal O_T$, the sets of patch indices of student/teacher that geometrically overlap with each other, and $m\colon \mathcal O_S \mapsto \mathcal O_T$, the function mapping the student indices to the spatially corresponding teacher patch.
Given student representation $r$ and teacher representation $h_k$ at layer $k$, the loss for a pair of elements at indices $i, j$ can be written as
\begin{equation}
\mathcal L_\text{struct}^{(i,j)}(r, h_k) = \left(\operatorname{cos}(g_k(r^{(i)}), g_k(r^{(j)})) - \operatorname{cos}(h^{(m(i))}, h^{(m(j))})\right)^2, \quad (i, j) \in \mathcal O_S \times \mathcal O_S,
\end{equation}
where $g_k$ is a learned non-linear projection for the $k$th teacher layer.
Thus, to compute the loss, only geometrically corresponding pairs of patches are used.
Given a set of target teacher layers $\mathcal K$, the total loss is then averaged over target layers and patch pairs: 
\begin{equation}
    \mathcal R_\text{struct}(x_S, x_T) = \frac{1}{|\mathcal K| |\mathcal O_S| (|\mathcal O_S|-1) } \sum_{k \in \mathcal K} \sum_{i \in \mathcal O_S} \sum_{\substack{j \in \mathcal O_S\\i\neq j}} \mathcal L_\text{struct}^{(i,j)}(r, h_k).
\end{equation}
We use target layers [0, 8, 16] for ViT-L (24 total layers), and [0, 10, 20, 30] for ViT-g and ViT-7B (40 total layers), where layer 0 refers to the output of the patch embedding.
The target features $h_k$ are processed by an element-wise layernorm without affine parameters before computing the cosine similarity.
We implement $g_k$ with a MLP of the same size as the projection head, except that it maps to an output space of dimensionality 1024 and does not have the final linear projection to logits.
The regularizer is applied both on global and local crops, and masked and unmasked patches.

Note that it is also possible to apply the regularizer only using student representations, and we successfully experimented with this option.
This simplifies the implementation, as patches at the same index do directly geometrically correspond to each other.
We opted for the slightly more complex teacher version as it allowed us to also apply the loss on the masked patches (which have no good similarity targets on the student), the EMA provides more stable targets, and the fact that the targets come from a different view may have an additional positive effect through enforcing a view invariance on the learned similarity structure.

\subsubsection{View Construction}
\label{app:sec:view-construction}

Given an image $x$, we sample student view $x_S$ and teacher view $x_T$ as follows.
We first extract a \emph{super crop} with pre-determined scale and random aspect ratio from the image $x$ and resize it to a square set of pixels with the side length being a multiple of the patch size.
The scale is randomly chosen such that an overlap of at least 10\% is achieved between the student and teacher crops when randomly picking the scale of the teacher crop from [0.32, 1.0] of the area of the original image.
Then, at random patch-grid aligned positions, we extract square student and teacher crops with a size matching the input resolution of the model.
A random horizontal flip is applied to the whole image before cropping with probability 0.5, such that all views share the same orientation.
Each crop is then independently transformed with standard photometric augmentations~\citep{grill2020bootstrap, caron2021emerging}, \ie color jittering, grayscale conversion, Gaussian blur, and solarization.
Different from DINO~\citep{caron2021emerging}, where the teacher processes the same augmented global crops as the student, teacher crops are augmented independently of their student crops.

For 80\% of the student crops, we additionally sample a patch mask $M$ that is applied to the student view, for which we use multi-block masking~\citep{zhou2021ibot,oquab2024dinov2} with cyclic shifts~\citep{darcet2025capi,venkataramanan2026franca} for the global crops, and uniform random masking for the local crops.
For global crops, we mask 65\% of the patches.
For local crops, we mask 50\% of the patches.

\subsubsection{Training Details}
\label{app:sec:training-details}

We base the implementations and hyperparameters of both \method and the DINOv2 baseline on the open source implementation of DINOv3~\citet{simeoni2025dinov3}.
Our self-trained DINOv2 baseline can thus be seen as a version of DINOv3 without gram anchoring and high-resolution refinement stages.

The complete training configurations are collected in \cref{app:sec:training-configurations}.
We give the backbone architectures and training hyperparameters shared by \methodNormal and the DINOv2 baselines in \cref{tab:shared-hyperparameters}, and their remaining training hyperparameters in \cref{tab:model-hyperparameters,tab:dinov2-baseline-hyperparameters}, respectively.
We note that the listed effective learning rate $\eta_\text{eff}$ is the result of square root scaling of a base learning rate $\eta_\text{base}$ with respect to a local batch size of 64: $\eta_\text{eff} = \sqrt{\frac{\text{global\_bs}}{64}} \cdot \eta_\text{base}$.

We summarize further experiment-specific settings in \cref{tab:model-hyperparameters,tab:ablation-settings}. 
In particular, the \method and DINOv2* ViT-L models listed in \cref{tab:in-results} are trained for 125k steps; the \method models shown in the ablation study are trained for 250k.
For the ``no EMA'' ablation, we set the teacher momentum to 0, effectively updating the teacher parameters with the current student parameters after each update step. 
We additionally change the configuration of the global collapse regularizer, increasing its overall weight relative to the alignment and structural collapse regularizers as well as increasing the weight of the marginal entropy component while decreasing the weight of the conditional entropy component.
We found this necessary to move the model out of the collapsed regime at the start of training.
Thus, removing the EMA teacher requires a slightly different hyperparameter regime due to the EMA's effect on the optimization dynamics, as can be expected.

The loss of \methodNormal is a weighted combination of (i) alignment loss (\cref{eqn:alignment-loss}) on visible ($\mathcal{L}_{\text{align}}^{\text{vis}}$) and masked ($\mathcal{L}_{\text{align}}^{\text{msk}}$) patches, (ii) global collapse regularization loss $\mathcal{R}_\text{global}$ (\cref{eq:global_collapse_regularization}) and (iii) structural regularization loss $\mathcal{R}_\text{struct}$ (\cref{eq:structural_regularization}):
\begin{equation}
    \mathcal{L} = w_\text{vis} \cdot \mathcal{L}_{\text{align}}^{\text{vis}} + w_\text{msk} \cdot \mathcal{L}_{\text{align}}^{\text{msk}}+ w_{\text{global}} \cdot \mathcal{R}_{\text{global}} + w_{\text{struct}} \cdot \mathcal{R}_\text{struct}.
\end{equation}
In practice, we use different loss coefficients for targets from different layers (see \cref{tab:model-hyperparameters}), and apply the structural regularizer only to the final layer representation.
All losses are applied to global and local crops.
Global and local crops are weighted according to their natural ratio by default.

\subsubsection{Data}
\label{app:sec:details-data}

Following the protocol established by \citet{oquab2024dinov2}, we construct \dataset{} by retrieving from a large pool images that align with the train splits of established public datasets, including IN-1k, IN-22k, ADE20K, etc.
Our source pool comprises approximately 750M images, obtained by deduplicating an internal collection of 1.3 billion images.
This process involves both self-deduplication which remove duplicates to improve retrieval quality, and relative-deduplication which prevents data contamination from the ``test'' splits of the seed datasets, which are used as standard benchmarks in evaluation.
Consistent with \citet{oquab2024dinov2}, we also implement NSFW filtering and facial blurring.

\subsection{Derivations}
\label{app:sec:derivations}

\begin{figure}[h]
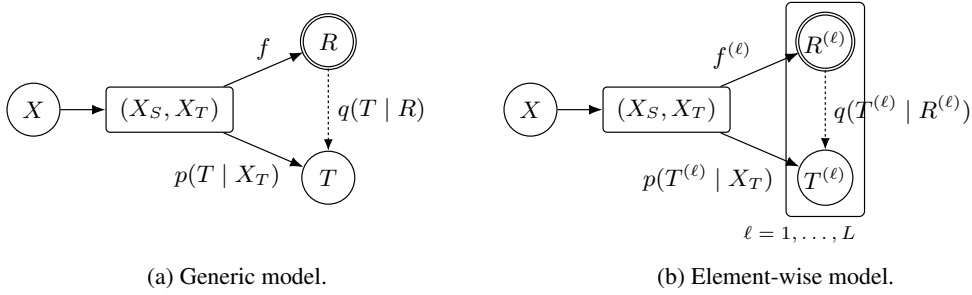

    \centering
    \begin{subfigure}[b]{.45\linewidth}
        \centering
        \includefigurepdf{figures/graphical-model-generic.pdf}
        \caption{Generic model.}
    \end{subfigure}\hfill
    \begin{subfigure}[b]{.53\linewidth}
        \centering
        \includefigurepdf{figures/graphical-model-elementwise.pdf}
        \caption{Element-wise model.}
    \end{subfigure}

    \caption{\textbf{Student-teacher graphical models.}
        (a) shows the generic model with the form of $T$ not specified; 
        (b) shows the element-wise model with additional factorization assumptions.
        The views $(X_S,X_T)$ are sampled jointly given input $X$.
        Solid arrows describe the sampling model, with $R=f(X_S)$ and
        $T\sim p(\cdot\mid X_T)$; double circles mark deterministic nodes.
        Dashed arrows denote the variational predictors $q(T\mid R)$ and
        $q(T^{(\ell)}\mid R^{(\ell)})$.
        In (b), $\ell$ indexes matched patch/layer pairs, with teacher targets
        relabeled accordingly and $R^{(\ell)}=f^{(\ell)}(X_S)$.
        The plate represents the factorizations
        $p(T\mid X_T)=\prod_{\ell=1}^L p(T^{(\ell)}\mid X_T)$ and
        $q(T\mid R)=\prod_{\ell=1}^L q(T^{(\ell)}\mid R^{(\ell)})$.
    }
    \label{fig:graphical-models}
\end{figure}

We assume $X_S$ and $X_T$ are jointly generated from $X$.
For the student, $R = f(X_S)$, where $f$ is a deterministic function, and thus $q(t \mid x_S):=q(t \mid f(x_S))$.
The teacher generates $T$ from $X_T$, resulting in the joint distribution $p(x_S,x_T,t)=p(x_S,x_T)\,p(t\mid x_T)$.
See \cref{fig:graphical-models} for graphical models.

\paragraph[Lower bound on mutual information]{Lower bound on mutual information (\cref{eqn:barber})}
This is the lower bound from \citet{barber2003algorithm}.
Starting with the definition of mutual information, we have:
\begin{align}
I(R; T)
    &= H(T) - H(T \mid R) \\
    &= H(T) + \mathbb E_{R,T} \log p(T \mid R) \\
    &= H(T) + \mathbb E_{R,T} \log q(T \mid R) + \mathbb E_{R,T} \log \frac{p(T \mid R)}{q(T \mid R)} \\
    &= H(T) + \mathbb E_{R,T} \log q(T \mid R) + \underbrace{\mathbb E_R \KLargs{p(\cdot \mid R)}{q(\cdot \mid R)}}_{>0} \\
    &\ge H(T) + \mathbb E_{X_S,T} \log q(T \mid X_S)\label{eqn:app:lower-bound}.
\end{align}

\paragraph[KL form of lower bound]{KL form of lower bound (\cref{eqn:lower-bound})}
Expanding the expected KL divergence:
\begin{align}
&\mathbb E_{X_S, X_T} \KLargs{p(\cdot \mid X_T)}{q(\cdot \mid X_S)} \\
    &\quad\quad = \mathbb E_{X_S, X_T} \mathbb E_{T \sim p(\cdot \mid X_T)} \left[\log p(T \mid X_T) - \log q(T \mid X_S)\right] \\
    &\quad\quad = \mathbb E_{X_T} \mathbb E_{T \sim p(\cdot \mid X_T)} \log p(T \mid X_T) - \mathbb E_{X_S} \mathbb E_{T \sim p(\cdot \mid X_T)}\log q(T \mid X_S)\\
    &\quad\quad = -H(T\mid X_T) - \mathbb E_{X_S, T} \log q(T\mid X_S).
\end{align}
It follows that $\mathbb E_{X_S, T} \log q(T\mid X_S) = -\KLargs{p(\cdot \mid X_T)}{q(\cdot \mid X_S)} - H(T\mid X_T)$.
Plugging this into \cref{eqn:barber} gives
\begin{align}
I(R; T) &\ge H(T) + \mathbb E_{X_S,T} \log q(T \mid X_S) \\
&= I(X_T; T) - \KLargs{p(\cdot \mid X_T)}{q(\cdot \mid X_S)}.
\end{align}

\paragraph[Element-wise decomposition of mutual information]{Element-wise decomposition of mutual information (\cref{eqn:anti-collapse-elementwise})}

With element-wise targets $T=(T^{(1)}, \ldots, T^{(L)})$, we have that
\begin{equation}
    H(T) = \sum_{\ell=1}^L H(T^{(\ell)}) - \TC(T),
\end{equation}
where $\TC(T) :=  \KLargs{(T^{(1)},\ldots,T^{(L)})}{\prod_{\ell=1}^{L}T^{(\ell)}}$ is the total correlation measuring dependence between the elements of $T$.
The total correlation becomes zero if and only if the variables are mutually independent; it is an extension of mutual information for more than two variables.
Then,
\begin{align}
I(X_T; T)
 &= H(T) - H(T \mid X_T) \nonumber\\
 &= \sum_{\ell=1}^L H(T^{(\ell)}) - \TC(T) - \left(\sum_{\ell=1}^L H(T^{(\ell)} \mid X_T) - \TC(T \mid X_T)\right) \nonumber\\
 &= \sum_{\ell=1}^L H(T^{(\ell)}) - \sum_{\ell=1}^L H(T^{(\ell)} \mid X_T) - \left(\TC(T) - \TC(T \mid X_T)\right) \nonumber\\
 &= \sum_{\ell=1}^L I(X_T; T^{(\ell)}) - \left(\TC(T) - \TC(T \mid X_T)\right).
\end{align}

\paragraph{Element-wise decomposition of KL divergence (\cref{subsec:alignemnt-loss})}
We want to show how the KL alignment term in \cref{eqn:lower-bound} decomposes for element-wise targets.
Let $T=(T^{(1)},\ldots,T^{(L)})$, and assume that the teacher
  and student distributions factorize conditionally over elements:
  \begin{equation}
      p(t\mid x_T) = \prod_{\ell=1}^L p(t^{(\ell)} \mid x_T),
      \qquad
      q(t\mid x_S) = \prod_{\ell=1}^L q(t^{(\ell)} \mid x_S).
  \end{equation}
  For any fixed pair of views $(x_S,x_T)$, we then have
  \begin{align}
      \KLargs{p(\cdot\mid x_T)}{q(\cdot\mid x_S)} &= \mathbb{E}_{T\sim p(\cdot\mid x_T)} \left[ \log \frac{p(T\mid x_T)}{q(T\mid x_S)} \right] \nonumber\\
      &=
      \mathbb{E}_{T\sim p(\cdot\mid x_T)}
      \left[
          \sum_{\ell=1}^L
          \log\frac{p(T^{(\ell)}\mid x_T)}
                    {q(T^{(\ell)}\mid x_S)}
      \right]
      \nonumber\\
      &=
      \sum_{\ell=1}^L
      \mathbb{E}_{T^{(\ell)}\sim p_\ell(\cdot\mid x_T)}
      \left[
          \log\frac{p(T^{(\ell)}\mid x_T)}
                    {q(T^{(\ell)}\mid x_S)}
      \right]
      \nonumber\\
      &=
      \sum_{\ell=1}^L \KLargs{p(\cdot\mid x_T)}{q(\cdot\mid x_S)}.
  \end{align}
  The third equality marginalizes out all elements other than
  $t^{(\ell)}$, since the logarithm depends only on that element.
  Taking the expectation over paired views gives
  \begin{equation}
      \mathbb{E}_{X_T,X_S}
      D_{\mathrm{KL}}\!\left(
          p(\cdot\mid X_T)\,\Vert\,q(\cdot\mid X_S)
      \right)
      =
      \sum_{\ell=1}^L
      \mathbb{E}_{X_T,X_S}
      \KLargs{p(\cdot\mid X_T)}{q(\cdot\mid X_S)}.
  \end{equation}

\subsection{Evaluation Details}
\label{app:sec:evaluation-details}

\subsubsection{Benchmark Protocols}
\label{app:sec:benchmark-protocols}

We evaluate models using standard lightweight probing setups.
Following \citet{simeoni2025dinov3}, we compare models of different patch sizes at matched input sequence length to equalize the compute of the forward pass.
All evaluations use final patch features of the backbone after the final layer norm.

\paragraph{Image Classification}
For classification, we apply the \emph{single-layer attention probing} setup from \citet{darcet2025capi}.
A learned query token is used to perform scaled multi-head attention (head dim=64) over key-value transformed features; the result is linearly projected to logits.
We perform grid search on a validation set to select the best hyperparameters.
Thereby, we scan the full cross product over learning rates from $(1 \cdot 10^{-5}, 2 \cdot 10^{-5}, 5 \cdot 10^{-5}, 1 \cdot 10^{-4}, 2 \cdot 10^{-4}, 5 \cdot 10^{-4}, 1 \cdot 10^{-3}, 2 \cdot 10^{-3}, 5 \cdot 10^{-3}, 1 \cdot 10^{-2})$ and weight decays from $(5 \cdot 10^{-4}, 1 \cdot 10^{-3}, 5 \cdot 10^{-2})$.
The probe is trained with a batch size of 1024 with AdamW for 12500 steps, with linear learning rate warm-up of 1250 steps followed by cosine decay.
The applied learning rate is linearly scaled with global batch as $\text{lr} = \frac{\text{global\_bs}}{256} \cdot \text{base\_lr}$.

For training on ImageNet, we split the original train set into training and validation sets, using 90\% for training, and 10\% for validation.
We apply light random augmentations during training, specifically RandomResizedCrop and random horizontal flips.
During evaluation, we resize the image to 256/292 pixels on the short side and take the central square crop of 224/256 pixels (87.5\% zoom crop) for patch size 14/16.
Models are evaluated on ImageNet val~\citep{russakovsky2015imagenet}. We report top-1 accuracy.

\paragraph{Semantic Segmentation}
Semantic segmentation is the task of assigning a class-label to every pixel of the input image. We use ADE20k \citep{zhou2019ade20k}, PASCAL VOC \citep{everingham2010pascal} and Cityscapes \citep{cordts2016cityscapes} datasets to evaluate this task and use the mean intersection-over-union (mIoU) as the reported metric.
We evaluate the task under two protocols with frozen backbones: kNN and linear probing. First, we use the kNN protocol for fast evaluations and assessment of the quality of dense patches. There, images and label maps are resized to $256\times256$ without cropping, using bicubic and nearest-neighbor interpolation respectively. The mIoU metric is computed on the full resized label maps.
Second, for the linear probing, we follow~\cite{simeoni2025dinov3} and take patch tokens from the last layer after the backbone's final LayerNorm.
On top of those, we train a linear head consisting of a BatchNorm layer followed by a $1\times1$ convolution that outputs $N$ logits, where $N$ is the number of classes in a given dataset.
The logits are bilinearly upsampled to the label resolution and trained with pixel-wise cross-entropy.
We scan the full cross product of learning rates \{$1\cdot10^{-4}$, $3\cdot 10^{-4}$, $1\cdot 10^{-3}$\} and weight decays \{$1\cdot 10^{-3}$, $1\cdot 10^{-2}$\}. All six probes are trained together on the same frozen features. Each probe is trained with AdamW optimizer (using $\beta_1=0.9$, $\beta_2 = 0.999$) and a batch size of 16 for 40,000 steps. The learning rate warms up linearly over 1,500 steps and then linearly decays to zero.
During training, we resize each image so that its short side is $448/512 \cdot s$ pixels for models with patch size 14/16, with $s$ drawn uniformly from [0.5, 2.0]. We then take a random $r\times r$ square crop adapted to 1024 tokens (i.e., $448^2/512^2$ crop for patch size 14/16) and apply random horizontal flips and photometric distortions (consisting of brightness, contrast, saturation and hue). During evaluation, we resize the image to 448/512 pixels on the short side and run $r\times r$ sliding-window inference with  windows and a stride of 298/341 pixels. Logits are averaged where windows overlap and bilinearly upsampled to the original image resolution.
We report the mIoU of the best probe over all hyperparameter configurations and evaluation steps.

\paragraph{Instance and Panoptic Segmentation}
Instance segmentation is the task of recognizing all pixels that belong to different object instances and their class-label. Panoptic segmentation~\citep{kirillov2019panoptic} extends instance segmentation beyond objects: every pixel is assigned a class label and, for countable objects (\emph{things}), an instance identity, while regions such as sky or road (\emph{stuff}) are only labeled by class.

We evaluate instance segmentation on the COCO dataset~\citep{lin2014coco}, and report the standard COCO average precision metric (mAP).
We evaluate panoptic segmentation on COCONut~\citep{deng2024coconut} and ADE20K-Panoptic~\citep{zhou2019ade20k}. For COCONut, we train on COCONut-S and evaluate on the relabeled COCO-val set with panoptic annotations; for ADE20K-Panoptic, we follow prior work~\citep{cheng2021per} and use the standard ADE20K training and validation splits. We report panoptic quality (PQ)~\citep{kirillov2019panoptic}, computed from IoU-based matching between predicted and ground-truth segments.

We use the same lightweight probe for both tasks, which operates on the last-layer patch features of the frozen backbone, taken after the final layer normalization.
The probe is split into two independent modules: a patch-wise linear classifier that predicts the semantic category of each patch, as in semantic segmentation, and a patch-wise 2-layer MLP that is used to split the patches into different segments (instances).
To get the masks, we compute the pairwise dot-product between the MLP outputs for different patches, whose sigmoid gives a score for the binary relation ``patches belong to the same segment''; we supervise the MLP with the binary cross entropy loss against patch pairs derived from the ground truth masks.
This is similar to the object probes used in \citet{li2025objectbinding}.

During training, images are augmented with random resized crops to $448\times 448$/$512\times 512$ pixels for patch size 14/16, i.e., square crops of $1024$ tokens; no other augmentation is used. Targets are defined at the patch level by taking the majority semantic and instance label within each patch. The semantic probe is trained with a cross-entropy loss, in which void patches are ignored, and a binary cross-entropy loss on the pairwise similarity predictions, with equal weights. We sweep learning rates over $\left\{ 1\cdot 10^{-4}, 3\cdot 10^{-4}, 1\cdot 10^{-3} \right\}$ and weight decays over $\left\{ 1\cdot 10^{-3}, 1\cdot 10^{-2} \right\}$, training all six probes jointly on the same frozen features. Each probe is trained with AdamW ($\beta_1 = 0.9$, $\beta_2 = 0.999$) and a batch size of 16 for 20 epochs, with a cosine learning-rate decay to $10^{-6}$ and no warm-up.

At inference, images are resized to the same square resolution without cropping. Each patch predicted as a non-void class $c$ yields a candidate soft mask, given by its pairwise similarities to all patches multiplied by their probabilities for class $c$. These candidate masks are filtered using class-wise mask NMS (IoU threshold 0.7), upsampled to the original image resolution, and binarized at 0.3; regions belonging to the same ``stuff'' class are merged. Metrics are computed at the original image resolution. Probes are evaluated after every epoch, and we select the best combination of learning rate, weight decay and epoch by validation mAP/PQ.

\paragraph{Depth Estimation}

Monocular depth estimation is the task of predicting a dense, per-pixel metric depth map from a single RGB image. It tests whether frozen features encode 3D scene geometry and not only semantics. We evaluate on two standard benchmarks: NYU Depth v2 \citep{silberman2012nyuv2}, which contains indoor scenes at $480\times 640$ resolution, and KITTI \citep{geiger2012kitti}, which contains outdoor driving scenes of about $375\times 1242$ pixels with sparse LiDAR ground truth. We report the root mean squared error (RMSE, in meters).
It is computed per image over valid pixels and averaged over the test set. Valid pixels are those whose ground truth lies strictly inside $[10^{-3}, d_{\text{max}}]$, with $d_{\text{max}} = 10$\,m on NYUv2 and $80$\,m on KITTI.

To predict monocular depth, we train a linear probe on top of the frozen backbone. The last-layer patch features, taken after the final layer normalization, are bilinearly upsampled by a factor of 4 and passed through a trained batch normalization layer and a $1\times 1$ convolution with 256 output channels.
To convert the outputs to depth, we predict it by classification over bins: the 256 outputs are passed through a ReLU, shifted by 0.1 and normalized into weights over 256 bins uniformly spaced in $[10^{-3}, d_{\text{max}}]$.
The predicted depth is the weighted sum of the bin centers.
During training, predictions are bilinearly upsampled to the label resolution and trained with the scale-invariant log loss~\citep{eigen2014depth}.
Similar to the linear semantic segmentation, we sweep learning rates over $\left\{ 1\cdot 10^{-4}, 3\cdot 10^{-4}, 1\cdot 10^{-3} \right\}$ and weight decays over $\left\{ 1\cdot 10^{-3}, 1\cdot 10^{-2} \right\}$, training all six probes jointly on the same frozen features.
Each probe is trained with AdamW and a batch size of 16 for 38,400 iterations, with a linear learning-rate warm-up over 12,800 iterations followed by cosine decay, with gradient norms clipped at 35.

Inputs are prepared in three steps: a fixed crop, an aspect-preserving resize, and then random square crops (training) or sliding windows (inference). On NYUv2, no fixed crop is applied: we use the full $480\times 640$ frame. On KITTI, we keep the bottom 352 rows and the central 1216 columns of each frame. We then resize the image so that its short side is $448/480$ pixels on NYUv2 and $308/352$ pixels on KITTI for patch size 14/16.
During training, we apply random rotations of up to $\pm 2.5^\circ$ and horizontal flips, each with probability $0.5$, jointly to the image and the depth map (resized with nearest-neighbor interpolation). We then take a random square crop whose side equals the resized short side ($448^2$/$480^2$ on NYUv2, $308^2$/$352^2$ on KITTI), and apply color augmentation (random gamma, brightness and per-channel color scaling) with probability $0.5$.

At inference, the whole resized image is processed with square sliding windows of the training crop size, spanning the full image height, with strides of 298/319 pixels on NYUv2 and 205/234 pixels on KITTI for patch size 14/16.
Bin logits are averaged where windows overlap, bilinearly upsampled to the ground-truth resolution, converted to depth, and averaged with the prediction for the horizontally flipped image. We report the configuration with the lowest RMSE, using the probe weights at the end of training.

\subsubsection{Baseline Details}
\label{app:sec:baseline-details}

\paragraph{Evaluated model checkpoints}
\Cref{app:tab:model-checkpoints} lists the external model checkpoints evaluated in this work.
Parameter counts refer to the evaluated backbone; training-only prediction heads and other auxiliary components are excluded.

\begin{table*}[h]
    \centering
    \caption{External checkpoints evaluated in \cref{tab:in-results}, grouped by method and ordered by first public appearance.}
    \label{app:tab:model-checkpoints}
    \scriptsize
    \setlength{\tabcolsep}{4pt}
    \renewcommand{\arraystretch}{1.08}
    \begin{tabular}{@{}>{\raggedright\arraybackslash}p{0.10\textwidth}>{\raggedright\arraybackslash}p{0.14\textwidth}>{\raggedright\arraybackslash}p{0.15\textwidth}lr>{\raggedright\arraybackslash}p{0.26\textwidth}@{}}
    \toprule
    Model & Citation & Arch. / Variant & Training Data & Params. & External Checkpoint \\
    \midrule
    MAE & \citet{he2021masked} & ViT-L/16 & IN-1k & 303M & \path{facebook/vit-mae-large} \\
    \addlinespace
    MSN & \citet{assran2022masked} & ViT-L/16 & IN-1k & 303M & \path{vitl16_600ep} \\
    \addlinespace
    data2vec 2.0 & \citet{baevski2023data2vec2} & ViT-L/16 & IN-1k & 303M & \path{large_imagenet} \\
    \addlinespace
    I-JEPA & \citet{assran2023self} & ViT-g/16 & IN-22k & 1,011M & \path{IN22K-vit.g.16-600e} \\
    \addlinespace
    DINOv2 (reg.) & \citet{oquab2024dinov2} & ViT-g/14 & LVD-142M & 1,136M & \path{dinov2_vitg14_reg} \\
    \addlinespace
    CAPI & \citet{darcet2025capi} & ViT-L/14 & IN-1k & 303M & \path{capi_vitl14_in1k} \\
    \addlinespace
    WebSSL & \citet{fan2025webssl} & ViT-7B/14 & MC-8B & 6,450M & \path{webssl_dino7b_full8b_224} \\
    \addlinespace
    Franca & \citet{venkataramanan2026franca} & ViT-g/14 (224 px; no RASA) & IN-22k & 1,135M & \path{franca_vitg14_In21K.pth} \\
    \addlinespace
    DINOv3 & \citet{simeoni2025dinov3} & ViT-7B/16 & LVD-1689M & 6,716M & \path{dinov3_vit7b16} \\
    \addlinespace
    V-JEPA 2.1 & \citet{murlabadia2026vjepa21} & ViT-G/16 & VM-163M & 1,843M & \path{vjepa2_1_vit_gigantic_384} \\
    \addlinespace
    LingBot-Vision & \citet{fu2026lingbotvision} & ViT-g/16 & Unk.-163M & 1,134M & \path{robbyant/lingbot-vision-vit-giant} \\
    \bottomrule
    \end{tabular}
\end{table*}

\paragraph{Estimation of number of training tokens}
\Cref{app:tab:training-tokens} details the training token estimates in \cref{tab:in-results}.
We count full student patch grids before masking, including all crops and mask variants, and exclude teacher and special tokens (\eg CLS, registers).
The total number of tokens in trillions is $T=10^{-12}\sum_r B_r S_r L_r$, where $B_r$, $S_r$, and $L_r$ are the batch size, optimizer update steps, and tokens per sample for each stage and input type.
For DINOv3's high resolution fine-tuning stage, which uses a mix of resolution, we use an expected 4,570 tokens per image.

We use training configurations and checkpoint metadata where available, and published recipes otherwise.
Note that this estimation is best-effort, as we cannot verify the actual training configurations from the released checkpoints.
MAE and MSN assume 1,600 and 600 training epochs, respectively; WebSSL assumes eight billion seen images.
I-JEPA uses only 44 completed IN-22k epochs, despite the checkpoint name \texttt{IN22K-vit.g.16-600e}, as indicated by the checkpoint metadata.
For V-JEPA~2.1, we combine batch size, number of steps, and resolution from the published schedule with the number of image and video masks from the example configurations as these values were not otherwise published.
V-JEPA video inputs also assume temporal tubelets of two frames.

\begin{table*}[p]
    \centering
    \caption{\textbf{Estimated training tokens for the checkpoints in \cref{tab:in-results}.}
    V-JEPA video inputs use temporal tubelets of two frames.
    Student views list crop counts and spatial dimensions in pixels, with frame counts for video.
    Batch size counts images or clips before crop and mask replication.
    Multi-stage models include a total row; total token counts are in trillions ($10^{12}$).
    Details and assumptions are given in \cref{app:sec:baseline-details}.}
    \label{app:tab:training-tokens}
    \footnotesize
    \setlength{\tabcolsep}{2pt}
    \renewcommand{\arraystretch}{1.12}
    \begin{tabular}{@{}lllrrlrr@{}}
    \toprule
    Model & \shortstack{Arch.} & Stage / input & \shortstack{Batch\\size} & Updates & Student views & \shortstack{Tokens /\\sample} & Tokens (T) \\
    \midrule
    MAE & L/16 & Pretraining & 4,096 & $\approx500{,}456$ & $1\times224^2$ & 196 & 0.402 \\ MSN & L/16 & Pretraining & 1,024 & $\approx750{,}684$ & $1\times224^2+10\times96^2$ & 556 & 0.427 \\ data2vec 2.0 & L/16 & Pretraining & 256 & 750,000 & $16\text{ masks}\times224^2$ & 3,136 & 0.602 \\ CAPI & L/14 & Pretraining & 16,384 & 500,000 & $1\times224^2$ & 256 & 2.097 \\ DINOv2* & L/16 & Pretraining & 2,048 & 125,000 & $2\times256^2+8\times112^2$ & 904 & 0.231 \\ \textbf{\methodNormal} & L/16 & Pretraining & 2,048 & 125,000 & $2\times256^2+8\times112^2$ & 904 & 0.231 \\ \midrule
    I-JEPA & g/16 & Pretraining & 2,048 & 304,920 & $1\times224^2$ & 196 & 0.122 \\ Franca & g/14 & Pretraining & 3,072 & 625,000 & $2\times224^2+8\times98^2$ & 904 & 1.736 \\ DINOv2* & g/16 & Pretraining & 3,072 & 500,000 & $2\times256^2+8\times112^2$ & 904 & 1.389 \\ \textbf{\methodNormal} & g/16 & Pretraining & 3,072 & 500,000 & $2\times256^2+8\times112^2$ & 904 & 1.389 \\ \addlinespace
    DINOv2 (reg.) & g/14 & Pretraining & 3,072 & 625,000 & $2\times224^2+8\times98^2$ & 904 & 1.736 \\ & & High-res. & 3,072 & 10,000 & $2\times518^2+8\times98^2$ & 3,130 & 0.096 \\ & & Total & & & & & \textbf{1.832} \\ \addlinespace
    LingBot-Vision & g/16 & Pretraining & 3,072 & 300,000 & $2\times256^2+8\times112^2$ & 904 & 0.833 \\ & & Gram anchoring & 3,072 & 100,000 & $2\times256^2+8\times112^2$ & 904 & 0.278 \\ & & High-res. & 3,072 & 100,000 & $2\times512^2+8\times112^2$ & 2,440 & 0.750 \\ & & Total & & & & & \textbf{1.860} \\ \midrule
    V-JEPA 2.1 & G/16 & Primary / image & 2,304 & 135,000 & $1\times256^2$ & 256 & 0.080 \\ & & Primary / video & 128 & 135,000 & $2\text{ masks}\times16\times256^2$ & 4,096 & 0.071 \\ & & Cooldown / image & 2,304 & 12,000 & $1\times512^2$ & 1,024 & 0.028 \\ & & Cooldown / video & 128 & 12,000 & $2\text{ masks}\times64\times384^2$ & 36,864 & 0.057 \\ & & Total & & & & & \textbf{0.235} \\ \midrule
    WebSSL & 7B/14 & Pretraining & 3,072 & $\approx2{,}604{,}167$ & $2\times224^2+8\times98^2$ & 904 & 7.232 \\ DINOv2* & 7B/16 & Pretraining & 4,096 & 500,000 & $2\times256^2+8\times112^2$ & 904 & 1.851 \\ \textbf{\methodNormal} & 7B/16 & Pretraining & 4,096 & 500,000 & $2\times256^2+8\times112^2$ & 904 & 1.851 \\ \addlinespace
    DINOv3 & 7B/16 & Pretraining & 4,096 & 1,000,000 & $2\times256^2+8\times112^2$ & 904 & 3.703 \\ & & Gram refinement & 4,096 & 70,000 & $2\times256^2+8\times112^2$ & 904 & 0.259 \\ & & High-res. & 2,048 & 10,000 & Mixed resolutions & 4,570 & 0.094 \\ & & Total & & & & & \textbf{4.056} \\ \bottomrule
    \end{tabular}
\end{table*}

\clearpage
\subsection{Training configurations}
\label{app:sec:training-configurations}

\begingroup
\raggedbottom
\begin{table}[H]
  \centering
  \caption{\textbf{Shared architectures and training hyperparameters for \methodNormal and DINOv2 baselines}. Entries spanning the three model columns are shared across all model sizes.}
  \label{tab:shared-hyperparameters}
  \scriptsize
  \setlength{\tabcolsep}{3pt}
  \begin{subtable}[t]{0.46\textwidth}
    \centering
    \caption{Backbone architectures.}
    \label{tab:backbone-architectures}
    \begin{tabular}[t]{@{}lccc@{}}
      \toprule
      \textbf{Hyperparameter} & \multicolumn{3}{c}{\textbf{Model size}} \\
      \cmidrule(lr){2-4}
      & \textbf{ViT-L} & \textbf{ViT-g} & \textbf{ViT-7B} \\
      \midrule
      Backbone parameters       & 303.1M & 1.13B & 6.7B \\
      Transformer blocks        & 24     & 40     & 40     \\
      Embedding dimension       & 1024   & 1536   & 4096   \\
      Attention heads           & 16     & 24     & 32     \\
      Attention-head dimension  & 64     & 64     & 128    \\
      FFN type                   & MLP    & SwiGLU & SwiGLU \\
      FFN hidden dimension      & 4096   & 4096   & 8192   \\
      Patch size                & \multicolumn{3}{c}{$16 \times 16$} \\
      Register tokens            & \multicolumn{3}{c}{4} \\
      Positional encoding       & \multicolumn{3}{c}{Axial RoPE} \\
      LayerScale initialization & \multicolumn{3}{c}{$1\!\times\!10^{-5}$} \\
      Drop-path rate            & 0.3    & 0.4    & 0.4    \\
      \bottomrule
    \end{tabular}
  \end{subtable}\hfill
  \begin{subtable}[t]{0.52\textwidth}
    \centering
    \caption{Shared training hyperparameters.}
    \label{tab:shared-training-hyperparameters}
    \begin{tabular}[t]{@{}lccc@{}}
      \toprule
      \textbf{Hyperparameter} & \multicolumn{3}{c}{\textbf{Model size}} \\
      \cmidrule(lr){2-4}
      & \textbf{ViT-L} & \textbf{ViT-g} & \textbf{ViT-7B} \\
      \midrule
      Training iterations       & \multicolumn{3}{c}{500k} \\
      Global batch size         & 2048   & 3072   & 4096   \\
      LR warm-up iterations     & \multicolumn{3}{c}{50k} \\
      Layer-wise LR decay       & 0.9    & 0.995  & 0.98   \\
      LR schedule after warm-up & Cosine decay & Constant & Constant \\
      Optimizer                 & \multicolumn{3}{c}{AdamW} \\
      AdamW $\beta_1$           & \multicolumn{3}{c}{0.9} \\
      AdamW $\beta_2$           & 0.999  & 0.99   & 0.99   \\
      Initial weight decay      & \multicolumn{3}{c}{0.04} \\
      Final weight decay        & 0.4    & 0.04   & 0.04   \\
      Weight decay schedule     & Cosine & Constant & Constant \\
      Teacher EMA momentum      & $0.992\!\xrightarrow{\text{cos}}\!1.0$ & 0.994 & 0.994 \\
      Global crops per sample   & \multicolumn{3}{c}{$2 \times 256^2$} \\
      Global crop weight        & \multicolumn{3}{c}{0.2} \\
      Local crops per sample    & \multicolumn{3}{c}{$8 \times 112^2$} \\
      Local crop weight         & \multicolumn{3}{c}{0.8} \\
      Parameter / reduction precision & \multicolumn{3}{c}{BF16 / FP32} \\
      \bottomrule
    \end{tabular}
  \end{subtable}
\end{table}

\begin{table}[H]
  \centering
  \caption{\textbf{\methodNormal training hyperparameters}.
  Entries spanning the three model columns are shared across all model sizes.
  Shared architectures and training hyperparameters are given in \cref{tab:shared-hyperparameters}. 
  Individual ablation settings are listed in \cref{tab:ablation-settings}.
  }
  \label{tab:model-hyperparameters}
  \scriptsize
  \setlength{\tabcolsep}{6pt}
  \begin{tabular}{@{}lccc@{}}
    \toprule
    \textbf{Hyperparameter} & \multicolumn{3}{c}{\textbf{Model size}} \\
    \cmidrule(lr){2-4}
    & \textbf{ViT-L} & \textbf{ViT-g} & \textbf{ViT-7B} \\
    \midrule
    \multicolumn{4}{@{}l@{}}{\textbf{Default training parameters}} \\
    Teacher-crop scale  & \multicolumn{3}{c}{[0.32, 1.0]} \\
    Number of categories           & \multicolumn{3}{c}{4096} \\
    Student and teacher temperature & \multicolumn{3}{c}{0.1} \\
    Global mask probability / patch ratio & \multicolumn{3}{c}{0.8 / 0.65} \\
    Local mask probability / patch ratio  & \multicolumn{3}{c}{0.8 / 0.50} \\
    Intermediate target blocks & \{8,16\} & \{10,20,30\} & \{10,20,30\} \\
    Struct.~regularizer target blocks & \{0,8,16\} & \{0,10,20,30\} & \{0,10,20,30\} \\
    Struct.~head output dimension & \multicolumn{3}{c}{1024} \\
    Proj./Struct hidden layers  & \multicolumn{3}{c}{3} \\
    Proj./Struct heads hidden dimension  & 2048 & 2048 & 4096 \\
    Proj. head bottleneck dimension  & 256 & 384 & 384 \\

    \multicolumn{4}{@{}l}{} \\
    Masked KL weight $w_\text{masked}$       & \multicolumn{3}{c}{0.8} \\
    Visible KL weight $w_\text{vis}$      & \multicolumn{3}{c}{0.2} \\
    Final layer weight             & \multicolumn{3}{c}{1} \\
    Early layer weight per-layer             & 0.25 & 0.166 & 0.166 \\
    Global collapse regularization weight $w_{\text{global}}$
    & \multicolumn{3}{c}{$1.0 \xrightarrow{\text{12.5k linear}} 0.1$} \\
    Structural regularizer weight $w_\text{struct}$ schedule
     & \multicolumn{3}{c}{0.0 $\xrightarrow{\text{12.5k linear}}$ 0.05 $\xrightarrow{\text{linear}}$ 1.0} \\
    
    \multicolumn{4}{@{}l}{} \\
    Base LR w.r.t. batch size 64 & $3 \times 10^{-4}$ & $1 \times 10^{-4}$ & $1.5 \times 10^{-4}$ \\
    Effective peak LR         & $1.697\!\times\!10^{-3}$ & $6.928\!\times\!10^{-4}$ & $1.200\!\times\!10^{-3}$ \\
    Effective final LR         & $5.697\!\times\!10^{-6}$ & $6.928\!\times\!10^{-4}$ & $1.200\!\times\!10^{-3}$ \\
    \midrule
    \multicolumn{4}{@{}l@{}}{\textbf{Ablation study on IN22k} (\cref{tab:component-ablations})} \\
    Training steps & 250k & & \\
    Base LR & $5 \times 10^{-4}$ & & \\
    \midrule
    \multicolumn{4}{@{}l@{}}{\textbf{ViT-L on IN1k} (\cref{tab:in-results})} \\
    Training steps & 125k & & \\
    Base LR & $1 \times 10^{-3}$ & & \\
    Global crop weight & 0.5 & & \\
    Local crop weight & 0.5 & & \\
    \bottomrule
  \end{tabular}
\end{table}

\begin{table}[H]
  \centering
  \caption{\textbf{Ablation settings.} Experiment-specific settings for the ablations in \cref{tab:component-ablations}. 
  All experiments use the shared ablation training settings in \cref{tab:model-hyperparameters}.
  }
  \label{tab:ablation-settings}
  \scriptsize
  \setlength{\tabcolsep}{2.5pt}
  \begin{subtable}[t]{0.32\textwidth}
    \vspace{0pt}
    \centering
    \caption{View construction (\cref{tab:ablation-view-construction}).}
    \label{tab:ablation-view-settings}
    \begin{tabular*}{\linewidth}{@{\extracolsep{\fill}}lr@{}}
      \toprule
      \textbf{Hyperparameter} & \textbf{Value} \\
      \midrule
      \multicolumn{2}{@{}l@{}}{\textbf{w/o~local crops}} \\
      Global crop weight & 1.0 \\
      Local crop weight & 0.0 \\
      \midrule
      \multicolumn{2}{@{}l@{}}{\textbf{w/o~masking}} \\
      Global mask probability & 0.0 \\
      Local mask probability & 0.0 \\
      \midrule
      \multicolumn{2}{@{}l@{}}{\textbf{w/o~stud.-teach. shift}} \\
      Min.~global crop S-T overlap & 1.0 \\
      \midrule
      \multicolumn{2}{@{}l@{}}{\textbf{w/o~independ. aug.}} \\
      Prob.~color jittering & 0.0 \\
      Prob.~grayscale & 0.0 \\
      Prob.~Gaussian blur & 0.0 \\
      Prob.~solarization & 0.0 \\
      \bottomrule
    \end{tabular*}
  \end{subtable}\hfill
  \begin{subtable}[t]{0.32\textwidth}
    \vspace{0pt}
    \centering
    \caption{KL alignment (\cref{tab:ablation-kl-composition}).}
    \label{tab:ablation-kl-settings}
    \begin{tabular*}{\linewidth}{@{\extracolsep{\fill}}lr@{}}
      \toprule
      \textbf{Hyperparameter} & \textbf{Value} \\
      \midrule
      \multicolumn{2}{@{}l@{}}{\textbf{w/o~multi-layer loss}} \\
      Early-layer weight per-layer & 0.0 \\
      \midrule
      \multicolumn{2}{@{}l@{}}{\textbf{w/o~visible-patch loss}} \\
      $w_\text{masked}$ & 1.0 \\
      $w_\text{vis}$ & 0.0 \\
      \midrule
      \multicolumn{2}{@{}l@{}}{\textbf{w/o~masked-patch loss}} \\
      $w_\text{masked}$ & 0.0 \\
      $w_\text{vis}$ & 1.0 \\
      \midrule
      \multicolumn{2}{@{}l@{}}{\textbf{w/o~EMA}} \\
      Teacher EMA momentum & 0.0 \\
      Marginal entropy weight & 1.5 \\
      Conditional entropy weight & 0.5 \\
      $w_{\text{global}}$ post-decay & 0.5 \\
      \bottomrule
    \end{tabular*}
  \end{subtable}\hfill
  \begin{subtable}[t]{0.32\textwidth}
    \vspace{0pt}
    \centering
    \caption{Objective (\cref{tab:ablation-loss-components}).}
    \label{tab:ablation-objective-settings}
    \begin{tabular*}{\linewidth}{@{\extracolsep{\fill}}lr@{}}
      \toprule
      \textbf{Hyperparameter} & \textbf{Value} \\
      \midrule
      \multicolumn{2}{@{}l@{}}{\textbf{w/o~$\mathcal{L}_{\text{align}}$}} \\
      $w_{\text{vis}}$ & 0.0 \\
      $w_{\text{masked}}$ & 0.0 \\
      \midrule
      \multicolumn{2}{@{}l@{}}{\textbf{w/o~$\mathcal{R}_{\text{global}}$}} \\
      $w_{\text{global}}$ & 0.0 \\
      \midrule
      \multicolumn{2}{@{}l@{}}{\textbf{w/o~$\mathcal{R}_{\text{struct}}$}} \\
      $w_{\text{struct}}$ & 0.0 \\
      \bottomrule
    \end{tabular*}
  \end{subtable}
\end{table}

\begin{table}[H]
\centering
\caption{\textbf{DINOv2 baseline training hyperparameters}.
Entries spanning all model columns are shared across model sizes.
Shared architectures and training hyperparameters are given in \cref{tab:shared-hyperparameters}. Loss weights are
configured coefficients.}
\label{tab:dinov2-baseline-hyperparameters}
\scriptsize
\setlength{\tabcolsep}{6pt}
\begin{tabular}{@{}lccc@{}}
  \toprule
  \textbf{Hyperparameter} & \multicolumn{3}{c}{\textbf{Model size}} \\
  \cmidrule(lr){2-4}
  & \textbf{ViT-L} & \textbf{ViT-g} & \textbf{ViT-7B} \\
  \midrule
  Global-crop scale         & \multicolumn{3}{c}{$[0.32,\,1.00]$} \\
  Local-crop scale          & \multicolumn{3}{c}{$[0.05,\,0.32]$} \\
  Global mask probability   & \multicolumn{3}{c}{0.5} \\
  Mask-ratio range          & \multicolumn{3}{c}{$[0.1,\,0.5]$} \\
  DINO prototypes           & 65{,}536 & 131{,}072 & 262{,}144 \\
  iBOT prototypes           & 65{,}536 & 131{,}072 & 98{,}304 \\
  DINO/iBOT heads layers    & \multicolumn{3}{c}{3} \\
  DINO head hidden dimension  & 2048 & 2048 & 8192 \\
  iBOT head hidden dimension  & 2048 & 2048 & 4096 \\
  DINO head bottleneck dimension & 256 & 384 & 512 \\
  iBOT head bottleneck dimension & 256 & 384 & 384 \\
  Student temperature       & \multicolumn{3}{c}{0.1} \\
  Teacher temperature       & \multicolumn{3}{c}{$0.04\!\rightarrow\!0.07$ over 50k iterations} \\

  \multicolumn{4}{@{}l}{} \\
  DINO                       & \multicolumn{3}{c}{1.0} \\
  iBOT                       & \multicolumn{3}{c}{0.5} \\
  KoLeo                      & \multicolumn{3}{c}{0.1} \\

  \multicolumn{4}{@{}l}{} \\
  Base LR w.r.t. batch size 64 & $3 \times 10^{-4}$ & $6 \times 10^{-5}$ & $6 \times 10^{-5}$ \\
  Effective peak LR
    & $1.697\!\times\!10^{-3}$
    & $4.157\!\times\!10^{-4}$
    & $4.800\!\times\!10^{-4}$ \\
  Effective final LR
    & $5.657\!\times\!10^{-6}$
    & $4.157\!\times\!10^{-4}$
    & $4.800\!\times\!10^{-4}$ \\
  Last-layer freeze         & \multicolumn{3}{c}{Initial 5k iterations} \\
  \bottomrule
\end{tabular}
\end{table}

\subsection{Qualitative Examples}
\label{app:sec:qualitative-examples}

We show visualizations of representations of a \method ViT-g model trained on \dataset in \cref{fig:pca-comparison} and \cref{fig:cosine-comparison}.
Input images have a resolution of $896 \times 896$ pixels, yielding a patch grid of $56 \times 56$ feature vectors.
PCA and cosine similarity visualizations are computed on the patch grid, and the resulting RGB image is upscaled to the input image size using nearest neighbor interpolation.

\clearpage
\endgroup

\newcommand{\featurepanel}[3]{\includegraphics[width=0.119\textwidth]{images/pca_cosine_viz/img_#1/#3.#2}}
\newcommand{\pcapair}[1]{\featurepanel{#1}{jpg}{original} & \featurepanel{#1}{png}{pca_rgb}}
\newcommand{\cosinepair}[1]{\featurepanel{#1}{jpg}{original} & \featurepanel{#1}{png}{cosine_center}}
\newcommand{\pcagridrow}[4]{\pcapair{#1} & \pcapair{#2} & \pcapair{#3} & \pcapair{#4} \\[-1pt]
}
\newcommand{\cosinegridrow}[4]{\cosinepair{#1} & \cosinepair{#2} & \cosinepair{#3} & \cosinepair{#4} \\[-1pt]
}

\begin{figure*}[p]
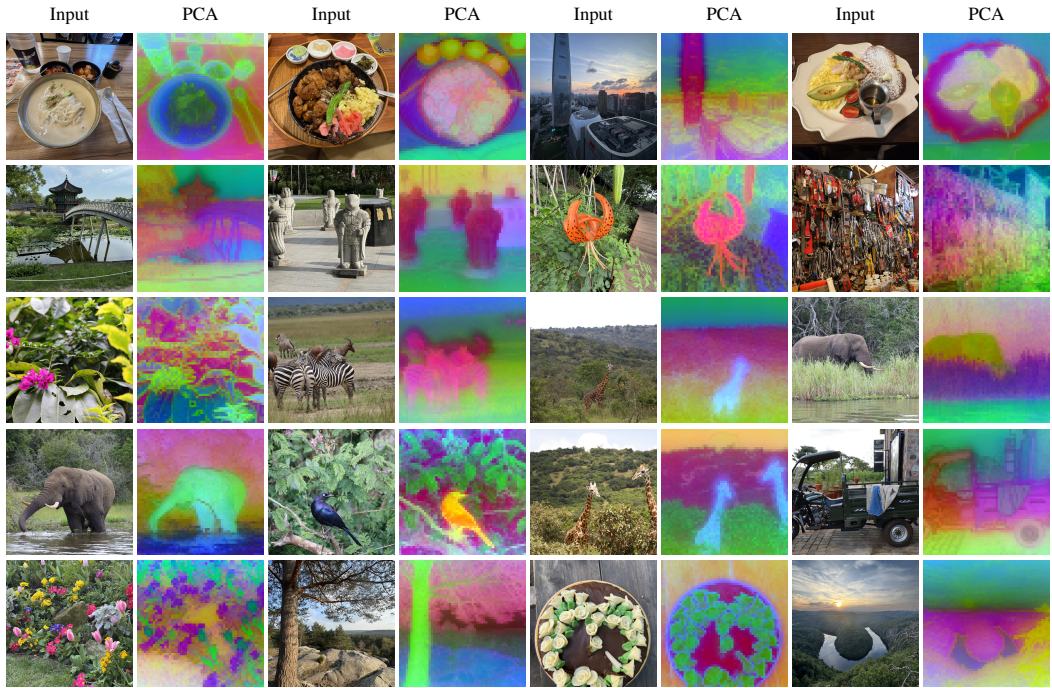

    \centering
    \setlength{\tabcolsep}{1pt}
    \begin{tabular}{@{}cccccccc@{}}
        \scriptsize Input & \scriptsize PCA &
        \scriptsize Input & \scriptsize PCA &
        \scriptsize Input & \scriptsize PCA &
        \scriptsize Input & \scriptsize PCA \\[2pt]
        \pcagridrow{0}{1}{2}{3}
        \pcagridrow{4}{5}{6}{7}
        \pcagridrow{8}{9}{10}{11}
        \pcagridrow{12}{13}{14}{15}
        \pcagridrow{16}{17}{18}{19}
    \end{tabular}
    \caption{\textbf{PCA visualization} of final-layer patch representations of a \method ViT-g model.
    For each of the 20 images, we show the input and its RGB projection onto the first three principal components computed individually on each image.
    }
    \label{fig:pca-comparison}
\end{figure*}

\begin{figure*}[p]
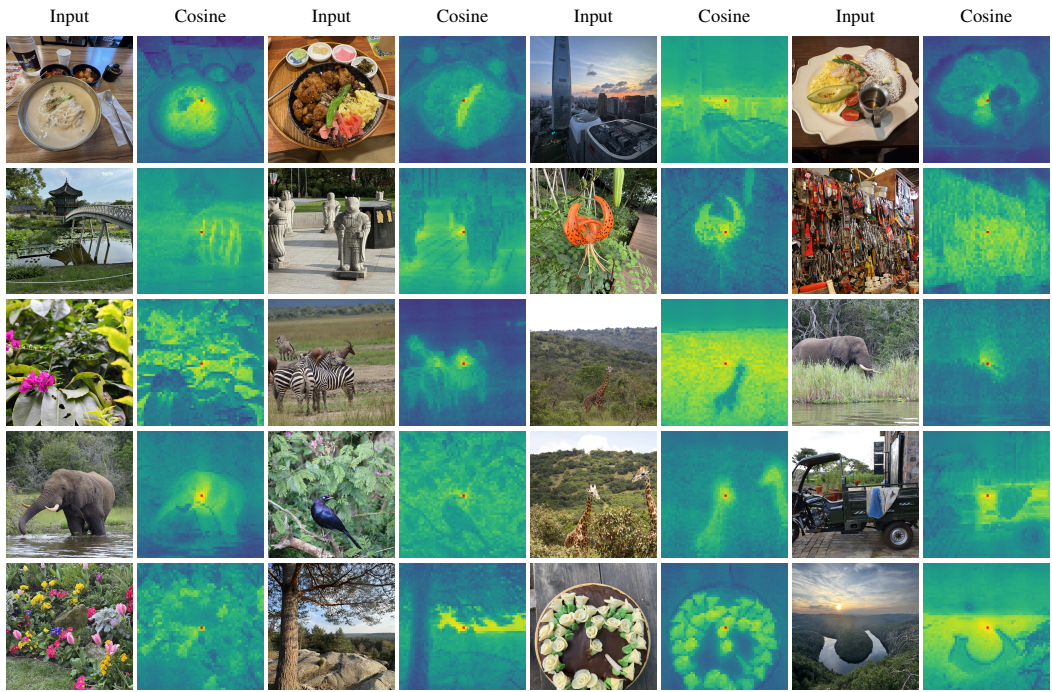

    \centering
    \setlength{\tabcolsep}{1pt}
    \begin{tabular}{@{}cccccccc@{}}
        \scriptsize Input & \scriptsize Cosine &
        \scriptsize Input & \scriptsize Cosine &
        \scriptsize Input & \scriptsize Cosine &
        \scriptsize Input & \scriptsize Cosine \\[2pt]
        \cosinegridrow{0}{1}{2}{3}
        \cosinegridrow{4}{5}{6}{7}
        \cosinegridrow{8}{9}{10}{11}
        \cosinegridrow{12}{13}{14}{15}
        \cosinegridrow{16}{17}{18}{19}
    \end{tabular}
    \caption{\textbf{Cosine similarity} between final-layer patch representations and the center-patch representation of a \method ViT-g model.
    Red markers identify the reference patch.
    The color scale interpolates between similarity values of 0 (full blue) and 1 (full yellow).
    }
    \label{fig:cosine-comparison}
\end{figure*}

\end{document}